%% file: main.tex
\pdfoutput=1
\RequirePackage{fix-cm}

\documentclass{svjour3}                     % onecolumn (standard format)
\smartqed  % flush right qed marks, e.g. at end of proof

\usepackage{natbib}
\setcitestyle{authoryear,open={(},close={)}}
\usepackage{subcaption}
\usepackage[utf8]{inputenc} % allow utf-8 input
\usepackage[T1]{fontenc}    % use 8-bit T1 fonts
\usepackage{hyperref}       % hyperlinks
\usepackage{url}            % simple URL typesetting
\usepackage{booktabs}       % professional-quality tables
\usepackage{amsfonts}       % blackboard math symbols
\usepackage{amsmath}
\usepackage{nicefrac}       % compact symbols for 1/2, etc.
\usepackage{microtype}      % microtypography
\usepackage{siunitx}
\usepackage{multirow}
\usepackage{breqn}
\usepackage[misc]{ifsym}
\usepackage{dsfont}
\usepackage{color}
\usepackage{tcolorbox}
\newcommand{\corrAuthor}{$^{\textrm{\Letter}}$}

\journalname{Journal}

\begin{document}
\title{Improving Human Decision-Making by Discovering Efficient Strategies for Hierarchical Planning}
\titlerunning{Discovering Efficient Strategies for Hierarchical Planning}        % if too long for running head

\author{Saksham Consul\textsuperscript{1} \and Lovis Heindrich\textsuperscript{1} \and Jugoslav Stojcheski\textsuperscript{1} \and Falk Lieder\textsuperscript{1, }\corrAuthor 
}

\authorrunning{S. Consul, L. Heindrich, J. Stojcheski, F. Lieder} % if too long for running head

\institute{\textsuperscript{1} Max Planck Institute for Intelligent Systems, T\"ubingen, Germany, 72076\\
            \corrAuthor  \email{falk.lieder@tuebingen.mpg.de}, ORCID: 0000-0003-2746-6110 \\
}
% \author{%
%   Saksham Consul \\
%   Max Planck Institute for Intelligent Systems\\
%   T\"ubingen, Germany, 72076\\
%   \texttt{saksham.consul@tuebingen.mpg.de} \\
%   \And
%   Lovis Heindrich \\
%   Max Planck Institute for Intelligent Systems\\
%   T\"ubingen, Germany, 72076\\
%   \texttt{lovis.heindrich@tuebingen.mpg.de} \\
%   \And
%   Jugoslav Stojcheski \\
%   Max Planck Institute for Intelligent Systems\\
%   T\"ubingen, Germany, 72076\\
%   \texttt{jugoslav.stojcheski@tuebingen.mpg.de} \\
%   \And
%   Falk Lieder \corrAuthor \\
%   Max Planck Institute for Intelligent Systems\\
%   T\"ubingen, Germany, 72076\\
%   \texttt{falk.lieder@tuebingen.mpg.de} \\
%   ORCID: 0000-0003-2746-6110 \\
% }
\date{Received: date / Accepted: date}

\sisetup{tight-spacing=true}
\maketitle
\begin{abstract}
%Big picture
To make good decisions in the real world people need efficient planning strategies because their computational resources are limited. 
%Specific issue
Knowing which planning strategies would work best for people in different situations would be very useful for understanding and improving human decision-making. But our ability to compute those strategies used to be limited to very small and very simple planning tasks. 
%Approach
To overcome this computational bottleneck, we introduce a cognitively-inspired reinforcement learning method that can overcome this limitation by exploiting the hierarchical structure of human behavior. The basic idea is to decompose sequential decision problems into two sub-problems: setting a goal and planning how to achieve it.
%Results
This hierarchical decomposition enables us to discover optimal strategies for human planning in larger and more complex tasks than was previously possible. The discovered strategies outperform existing planning algorithms and achieve a super-human level of computational efficiency. 
We demonstrate that teaching people to use those strategies significantly improves their performance in sequential decision-making tasks that require planning up to eight steps ahead. By contrast, none of the previous approaches was able to improve human performance on these problems.
%Conclusion
These findings suggest that our cognitively-informed approach makes it possible to leverage reinforcement learning to improve human decision-making in complex sequential decision-problems. Future work can leverage our method to develop decision support systems that improve human decision making in the real world. \\

\textbf{Keywords}: decision-making; planning; automatic strategy discovery; reinforcement learning; resource rationality; boosting\\

{\small \textbf{Acknowledgements: }
This project was funded by grant number CyVy-RF-2019-02 from the Cyber Valley Research Fund. The authors would like to thank Yash Rah Jain, Frederic Becker, Aashay Mehta, and Julian Skirzynski for helpful discussions.\\}

\end{abstract}

\section{Introduction}
To make good decisions people often plan many steps ahead. This requires efficient planning strategies because the number of possible action sequences grows exponentially with the number of steps and people's cognitive resources are limited. Recent work has shown that teaching people clever decision strategies is a promising way to improve human decision-making \citep{Hertwig2017,Hafenbradl2016}; this approach is known as \textit{boosting}. One of the bottlenecks of boosting is that discovering clever decision strategies that work well in the real world is very challenging and time consuming. For boosting to be effective the taught strategies have to be well-adapted to the decisions and environments in which people will use them \citep{Simon1956,Gigerenzer2002,Todd2012}.
The cognitive modeling paradigm of resource-rational analysis \citep{LiederGriffiths2020} can be used to mathematically define planning strategies that are optimally adapted to the problems people have to solve and the cognitive resources people can use to solve those problems \citep{CallawayLiederEtAl2018, Callaway2021}.
%Specific issue
Knowing those strategies can be very useful for understanding and improving human decision-making \citep{CallawayLiederEtAl2018, Callaway2021, CognitiveTutorsRLDM, CognitiveTutorsPNAS}. Recent work has developed algorithms for computing such optimal strategies from a model of the problem to be solved, the cognitive operations people have available to solve that problem, and how costly those operations are \citep{CallawayLiederEtAl2018, Callaway2021, lieder2017automatic, Griffiths2020}. We refer to this approach as \textit{automatic strategy discovery}. This approach frames planning strategies as policies for selecting planning operations. Its methods use algorithms from dynamic programming and reinforcement learning \citep{SuttonBarto} to compute the policy that maximizes the expected reward of executing the resulting plan minus the cost of the computations that the policy would perform to arrive at that plan \citep{callaway2017learning,LiederGriffiths2020,Griffiths2019}. Recent work used dynamic programming to discover optimal planning strategies for different three-step planning problems, and found that it is possible to improve human planning on those problems by teaching people the automatically discovered strategies \citep{CognitiveTutorsRLDM,CognitiveTutorsPNAS}. Subsequent work applied reinforcement learning to approximate strategies for planning up to six steps ahead in a task where each step entailed choosing between two options and there were only two possible rewards \citep{callaway2017learning}. But 
% The planning discovery algorithms 
none of the existing strategy discovery methods \citep{callaway2017learning, KemturJain2020}
% plan by considering all sequential steps in a single stage. This severely limits the scalability
is scalable enough to discover good planning strategies for more complex environments.
% with longer sequences
%due to the exponentially-increasing number of possible states and their applicability is strongly limited to tiny environments where extensive computations are tractable.
% This is 
This is because the run time of these methods grows exponentially with the size of the planning problem. This confined automatic strategy discovery methods to very small and very simple planning tasks. Discovering planning strategies that achieve -- let alone exceed -- the computational efficiency of human planning is still out of reach for virtually all practically relevant sequential decision problems.

%Approach
%Furthermore, recent work in cognitive science has found that human planning resembles the sequence of planning operations one would arrive at by optimal metareasoning about how to maximize the expected improvement in decision quality minus the cost of computation \cite{callaway2017learning,LiederGriffiths2020,Griffiths2019}. More generally, \textit{planning strategies} can be thought of as policies that select planning operations. The optimal planning strategy 
% is the one that achieves the best trade-off between the expected return of the resulting plan and the cost of the planning operations it performed to generate the plan. Using AI and machine learning to compute such strategies is known as \textit{automatic strategy discovery}. 
To overcome this computational bottleneck, we developed a scalable % strategy discovery 
 method for
% that can be used to 
discovering planning strategies that achieve a (super-)human level of computational efficiency on some of the planning problems that are too large for existing strategy discovery methods. % which further limits the scalability of such strategy discovery methods.
Our approach draws inspiration from the hierarchical structure of human behavior \citep{Botvinick2008, Miller1960, carver2001self, Tomov2020}. Research in cognitive science and neuroscience suggests that the brain decomposes long-term planning into goal-setting and planning at multiple hierarchically-nested timescales \citep{carver2001self,Botvinick2008}.
%Standard methods for solving sequential decision problems face issues when planning for long horizons due to the huge number of possible states and uncertainty for actions far in the future.
% This computational efficiency enables people to plan many steps into the future with very few computations. 
% While the exact mechanism of human planning remain unknown, there is evidence that the brain decomposes long-term planning into goal-setting and planning at multiple hierarchically nested timescales \cite{carver2001self,Botvinick2008}. %\cite{carver2001self, botvinick2014model} discuss the various cognitive advantages of breaking a goal to multiple sub-goals and planning for each sub-goal. Advantages infamaclude multiple sub-goal combinations to a particular goal, utilization of pre-learned action for a sub-goal (habits), etc.
% According to \cite{carver2001self} gives evidence about the hierarchy of goals in human planning and how it can be decomposed into multiple abstract sub-goal layers. Additionally, it establishes that a goal at any given level can be achieved in a variety of means in the lower level. Furthermore, \cite{botvinick2014model} discusses various quantitative results supporting the hierarchical setting in human planning.
Furthermore, \citet{solway2014optimal} found that human learners spontaneously discover optimal action hierarchies. 
Inspired by these findings, we extend the near-optimal strategy discovery method proposed in \citet{callaway2017learning} by incorporating hierarchical structure into the space of possible planning strategies. Concretely, the planning task is decomposed into first selecting one of the possible final destinations as a goal solely based on its own value and then planning the path to this selected goal.

%Results and Significance
We find that imposing hierarchical structure makes automatic strategy discovery methods significantly less computationally expensive without compromising the performance of the discovered strategies. Our hierarchical decomposition leads to a substantial reduction in the computational complexity of the strategy discovery problem, which 
% manifests in substantial improvements even for small problem instances and 
makes it possible to scale up automatic strategy discovery to many planning problems that were prohibitively large for previous strategy discovery methods. 
% Depending on the structure of the environment, our approach increases the scalability of automatic algorithm discovery by a factor of $5^{2484}$. 
This allowed our method to discover planning strategies that achieve a super-human level of computational efficiency on non-trivial planning problems. We demonstrate that this advance makes it possible to improve human decision-making in larger and more complex planning tasks than was previously possible. %In consequence, we can
The plan for this article is as follows: We start by introducing the frameworks and methods that our approach builds on. %Next, we present evidence of hierarchical planning in humans and need for hierarchical strategy discovery. 
We then present our new reinforcement learning method for discovering hierarchical planning strategies. Next, we evaluate our method's performance and scalability against the state of the art. We then test whether the resulting advances are sufficient to improve human decision making in complex planning problems and close by discussing the implications of our findings for the development of more intelligent agents, understanding human planning \citep{LiederGriffiths2020}, and improving human decision-making \citep{CognitiveTutorsRLDM}.

\section{Background and related work}

Before we introduce, evaluate, and apply our new method for discovering hierarchical planning strategies we now briefly introduce the concepts and methods that it builds on. We start by introducing the theoretical framework we use to define what constitutes a good planning strategy.

% In this section, we briefly review all previous automatic planning discovery methods utilising value of information (VOI) and the paradigm utilised throughout the paper.

% \subsection{Discovering Optimal Planning Strategies By Solving Metalevel MDPs}
% \js{I think we should move this paragraph to the ``The Mouselab-MDP paradigm'' section below in order to avoid having two defintions of a metalevel MDP...}

% \cite{CallawayLiederEtAl2018} developed a method to automatically discover optimal planning methods by modeling the optimal planning strategy as a solution to a meta-level Markov Decision Process (metalevel MDP). A metalevel MDP $\mathcal{M} = (\mathcal{B}, \mathcal{C}, \mathcal{T}, \mathcal{R})$ is defined as an undiscounted MDP where the states $\mathcal{B}$ encode the agent's beliefs, $\mathcal{T}(b,c,b')$ is the probability of transitioning from belief state $b$ to belief state $b'$ by performing the computation $c \in \mathcal{C}$, and $\mathcal{R}$ is a reward function that describes the costs and benefits of computation \cite{hay2014selecting}. It is important to note that the actions in a metalevel MDP are computations which are different from object level actions in which the agent interacts physically with the environment.

\subsection{Resource rationality} \label{sec:RR}

Previous work has shown that people's planning strategies are jointly shaped by the structure of the environment and the cost of planning \citep{CallawayLiederEtAl2018,Callaway2021}. This idea has been formalized within the framework of resource-rational analysis \citep{LiederGriffiths2020}. Resource-rational analysis is cognitive modeling paradigm that derives process models of people's cognitive strategies from the assumption that the brain makes optimal use of its finite computational resources. These computational resources are model as a set of elementary information processing operations. Each of these operations has a cost that reflects how much computational resources it requires. Those operations are assumed to be the building blocks of people's cognitive strategies. To be resource-rational a planning strategy has to achieve the optimal tradeoff between the expected return of the resulting decision and the expected cost of the planning operation it will perform to reach that decision. Both depend on the structure of the environment. The degree to which a planning strategy ($h$) is resource-rational in a given environment ($e$) can be quantified by the the sum of expected rewards achieved by executing the plan it generates ($R_{\text{total}}$) minus the expected computational cost it incurs to make those choices, that is
\begin{dmath}
    \text{RR}(h,e)=\mathds{E}\left[ R_{\text{total}} |\; h,e \right]
    - \lambda \cdot \mathds{E}\left[ N |\; h,e \right], \label{eq:RR}
\end{dmath}
where $\lambda$ is the cost of performing one planning operation and $N$ is the number of planning operations that the strategy performs. Throughout this article, we use this measure as our primary criterion for the performance of planning algorithms, automatically discovered strategies, and people.

\subsection{Discovering resource-rational planning strategies by solving metalevel MDPs}
\citet{CallawayLiederEtAl2018} developed a method to automatically derive resource-rational planning strategies by modeling the optimal planning strategy as a solution to a metalevel Markov Decision Process (metalevel MDP). In general, a metalevel MDP 
$M = (\mathcal{B}, \mathcal{C}, T, r)$
is defined as an undiscounted MDP where $b \in \mathcal{B}$ represents the belief state, $T(b,c,b')$ is the probability of transitioning from belief state $b$ to belief state $b'$ by performing computation $c \in \mathcal{C}$, and $r(b, c)$ is a reward function that describes the costs and benefits of computation \citep{hay2014selecting}. 
It is important to note that the actions in a metalevel MDP are computations which are different from object-level actions -- the former are planning operations and the latter are physical actions that move the agent through the environment. Previous methods for discovering near-optimal decision strategies \citep{lieder2017automatic,CallawayLiederEtAl2018,callaway2017learning} have been developed for and evaluated in a planning task known as the Mouselab-MDP paradigm \citep{CallawayLiederKrueger2017,Callaway2021}. 

\subsection{The Mouselab-MDP paradigm}  \label{sec:MouselabMDP}

%\js{I guess we should break this paragraph into 2-3 smaller ones.}

\begin{figure}[ht]
    \centering
    \includegraphics[width=0.75\textwidth]{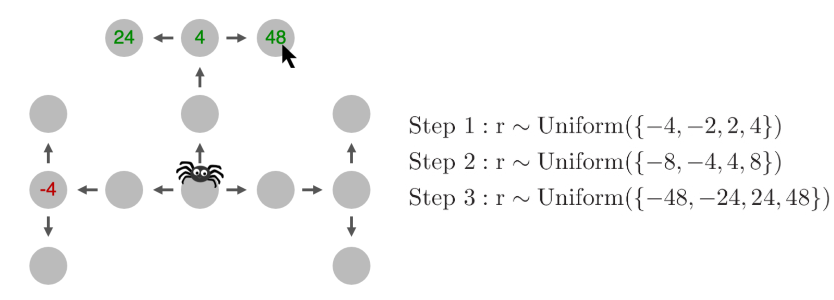}
    \caption{Illustration of the Mouselab-MDP paradigm. Rewards are revealed by clicking with the mouse, prior to selecting a path using the keyboard. This figure shows one concrete task that can be created using this paradigm. Many other tasks can be created by varying the size and layout of the environment, the distributions that the rewards are drawn from, and the cost of clicking.
    %\fl{It is nice that the image is so sharp. The text should not be in italics though.}
    }
    \label{fig:MouselabMDPParadigm}
\end{figure}

The Mouselab-MDP paradigm was developed to make people's elementary planning operations observable. This is achieved by externalizing the process of planning as information seeking. Concretely, the Mouselab-MDP paradigm illustrated in Figure~\ref{fig:MouselabMDPParadigm} shows the participant a map of an environment where each location harbors an occluded positive or negative reward. To find out which path to take the participant has to click on the locations they consider visiting to uncover their rewards. Each of these clicks is recorded and interpreted as the reflection of one elementary planning planning operation. The cost of planning is externalized by the fee that people have to pay for each click. People can stop planning and start navigating through the environment at any time. But once they have started to move through the environment they cannot resume planning.
% until it reaches one of the goal nodes.
% and reach one of the goal nodes (represented by the dark red and green nodes).
The participant has to follow one of the paths along the arrows to one of the outermost nodes. %Its goal is to maximize the sum of the rewards along the chosen path minus the total cost of the planning operations it performed to select that path.
% The planning strategy is the sequence of computations performed by the agent,

To evaluate the resource-rational performance metric specified in Equation~\ref{eq:RR} in the Mouselab-MDP paradigm, we measure $R_{\text{total}}$ by the sum of rewards along the chosen path, set $\lambda$ to the cost of clicking, measure $N$ by the number of clicks that a strategy made on a given trial. %In the Mouselab-MDP environments used in this article, the cost of clicking is always one (i.e., $\lambda=1$).

The structure of a Mouselab-MDP environment can be modelled as a directed acyclic graph (DAG), where each node is associated with a reward that is sampled from a probability distribution, and each edge represents a transition from one node to another. In this article, we refer to the agent's initial position as the \textit{root node}, the most distant nodes as \textit{goal nodes} and all other nodes as \textit{intermediate nodes}.

\begin{figure}[ht]
  \centering
  \includegraphics[height=0.35\textwidth]{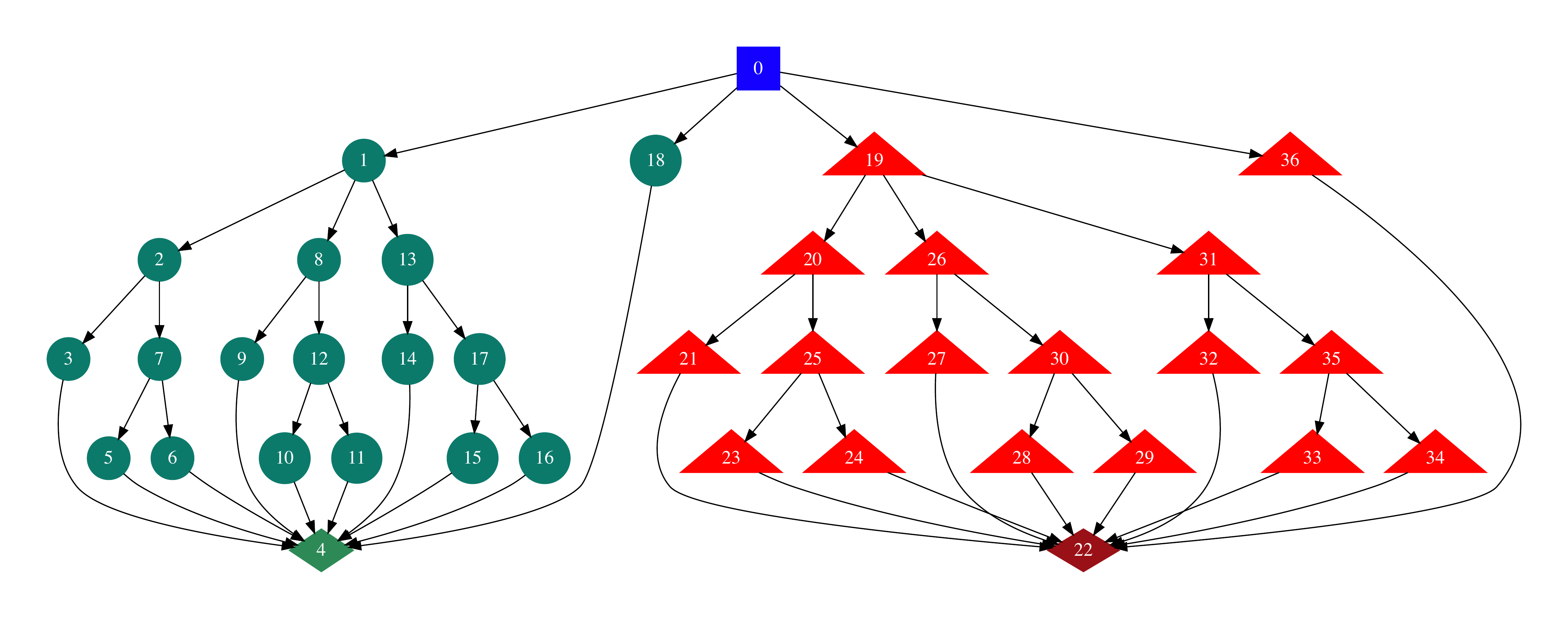}
  \caption{Mouselab-MDP environment with 2 goals. Nodes associated with
  each goal are denoted in green (circles) and red (triangles), respectively. The goal nodes have darker shades of green and red (diamonds), and the root node's color is blue (square).}
  \label{fig:env}
\end{figure}

Figure~\ref{fig:env} shows an instance of a Mouselab-MDP environment that we use extensively in this article. There, the variance of each node's reward distribution increases with the node's depth\footnote{A node's depth is defined as the length of the longest path connecting this node to the root node.}. This models that the values of distant states are more variable than the values of proximal states. Therefore, the goal nodes have a higher variance than the intermediate nodes.

\subsection{Resource-rational planning in the Mouselab-MDP paradigm}
To discover optimal planning strategies, we can draw on previous work that formalized resource-rational planning in the Mouselab-MDP paradigm as the solution to a metalevel MDP \citep{CallawayLiederEtAl2018,Callaway2021}. 
% As discussed, automatic planning strategy discovery can be achieved by solving metalevel MDPs. 
%Since planning is an internal cognitive process, it is impossible to observe the planning process directly. Therefore, it has to be inferred using process-tracing paradigms such as the Mouselab-MDP paradigm \cite{CallawayLiederKrueger2017}, which was developed to study such processes.
% The Mouselab-MDP paradigm was developed to study the discovered planning stategies \cite{CallawayLiederKrueger2017}. 
 %The goal nodes are the terminal nodes in the DAG (i.e., the diamonds in Figure~\ref{fig:env}). 
%Initially, the agent is located at the root node and the reward of each node is concealed. 
% The clicks that people make in the Mouselab-MDP paradigm correspond to a series of computations, each of which reveals the value of a specific node and has a certain cost (trading computational resources for information).
% Following the strategy discovery method by \cite{lieder2017automatic}, we model finding optimal planning strategy
% % with normally-distributed rewards
% as a solution to a 
In the corresponding metalevel MDP each belief state $b \in \mathcal{B}$ encodes
% a multivariate normal distributions 
probability distributions over the rewards that the nodes might harbor. The possible computations are $\mathcal{C} = \{ \xi_1, ..., \xi_M, c_{1,1}, ..., c_{M,N}, \perp \}$, where $c_{g,n}$ reveals the reward at intermediate node $n$ on the path to goal $g$, and $\xi_g$ reveals the value of the goal node $g$. For simplicity, we set the cost of each computation to $1$. When the value of a node is revealed, the new belief about the value of the inspected node assigns a probability of one to the observed value.
%of goal set $h_{g} \in \mathcal{H}$. 
% \red{Furthermore, a goal set $h_{g} \in \mathcal{H}$ refers to all nodes which lie on all paths leading to a goal $g$ (including the goal node itself)}
The metalevel operation
$\perp$ terminates planning and triggers the execution of the plan. The agent selects one of the paths to a goal state that has the highest expected sum of rewards according to the current belief state. %Whatever path the agent chooses has to end in one of the goal states.
%\lovis{Lot of similar information as in 2.1}
%The trade-off between the cost and the value of collecting information
% and the value of collected information
%is necessary to establish a non-trivial solution.

% \section{The need of hierarchical planning strategies}
% \js{I think we should move the part after this comment to the ``Introduction'' section. If this suggestion is accepted, then we can merge (what is left in) this section with the next section ``Discovering Hierarchical Planning Strategies''.}

\subsection{Methods for solving meta-level MDPs}\label{sec:metaMDP}
In their seminal paper, 
\citet{russell1991right} introduced the theory of rational metareasoning. In \citet{russell1992principles}, they define the value of computation $\mathrm{VOC}(c,b)$ to be the expected improvement in decision quality achieved by performing computation $c$ in belief state $b$ and continuing optimally, minus the cost of computation $c$. Using this formalization, the optimal planning strategy $\pi_{\mathrm{meta}}^{*}$ is a selection of computations which maximizes the value of computation (VOC), that is
\begin{equation}
    \pi_{\mathrm{meta}}^{*} = \mathrm{arg}\max_c \mathrm{VOC}(c,b).
\end{equation}
When the VOC is non-positive for all \textit{available} computations, the policy terminates ($c = \perp$) and executes the best object-level action according to the current belief state. Hence, $\mathrm{VOC}(\perp, b) = 0$.
In general, the VOC is computationally intractable but it can be approximated \citep{callaway2017learning}. % and various attempts tried to approximate VOC in order to find near-optimal strategies.
\citet{lin2015metareasoning} estimated VOC by the myopic value of computation ($\mathrm{VOI}_{1}$), which is the expected improvement in decision quality that would be attained by terminating deliberation immediately after performing the computation. %Their method selects in which a computation is selected greedily as it is the last computation before termination of planning. 
\citet{hay2014selecting} approximated rational metareasoning by solving multiple smaller metalevel MDPs that each define the problem of deciding between one object-level action and its best alternative. 

\subsubsection{Bayesian Metalevel Policy Search}
Inspired by research on how people learn how to plan \citet{KruegerLieder2017}, \citet{callaway2017learning} developed a reinforcement learning method for learning when to select which computation. This method uses Bayesian optimization to find a policy that maximizes the expected return of a metalevel MDP. The policy space is parameterized by weights that determine to which extent computations are selected based on the myopic VOC versus less short-sighted approximations of the value of computation. It thereby improves upon approximating the value of computation by the myopic VOC by considering the possibility that the optimal metalevel policy might perform additional computations afterwards. Concretely, BMPS approximates the value of computation by interpolating between the myopic VOI and the value of perfect information, that is 
\begin{equation}
\label{eqn:bmps}
    \mathrm{\widehat{VOC}}(c,b; \mathbf{w}) = w_{1} \cdot \mathrm{VOI}_{1}(c,b) +  w_{2} \cdot \mathrm{VPI}(b) + w_{3} \cdot \mathrm{VPI_{sub}}(c,b) - w_{4} \cdot \mathrm{cost}(c),
\end{equation}
where $\mathrm{VPI}(b)$ denotes the value of perfect information. $\mathrm{VPI}(b)$ assumes that all computations possible at a given belief state would take place. Furthermore, $\mathrm{VPI_{sub}}(c,b)$ measures the benefit of having full information about the subset of parameters that the computation reasons about (e.g., the values of all paths that pass through the node evaluated by the computation), $\mathrm{cost}(c)$ is the cost of the computation $c$, and $\mathbf{w} = (w_{1}, w_{2}, w_{3}, w_{4})$ is a vector of weights. Since the VOC and VPI$_{\mathrm{sub}}$ are bounded by the VOI$_{1}$ from below and by the VPI from above, the approximation of VOC (i.e. $\mathrm{\widehat{VOC}}$) is a convex combination of these features, and the weights associated with these features are constrained to a probability simplex set. Finally, the weight associated with the cost function $w_{4} \in [1, h]$, where $h$ is the maximum number of available computations to be performed. The value of these weights are computed using Bayesian Optimization \citep{mockus2012bayesian}. Discovery of the optimized weights, is analogous to discovering the optimal policy in the environment.

\subsubsection{Alternative approaches}
Alternative methods to solve metalevel MDPs include works by \citep{sezener2020static} and \citep{svegliato2018adaptive}. \citet{sezener2020static} solves a multi-arm bandit problem using a Monte Carlo Tree Search based on static and dynamic value of computations. In a bandit problem, unlike most models of planning, transitions depend purely on the chosen action and not on the current state. \citet{svegliato2018adaptive} devised an approximate metareasoning algorithm using temporal difference (TD) learning to decide when to terminate the planning process. %It does not discuss how metalevel computations in the planning stage should be selected.

\subsection{Intelligent cognitive tutors} \label{sec:cogtutbackground}
Utilising the optimal planning strategies discovered by solving metalevel MDPs, \citet{CognitiveTutorsRLDM, CognitiveTutorsPNAS} have developed intelligent tutors that teach people the optimal planning strategies for a given environment. Most of the tutors let people practice planning in the Mouselab-MDP paradigm and provide them immediate feedback on each chosen planning operation. The feedback is given in two ways: (1) information about what the optimal planning strategy would have done; and (2) an affective element given as a positive feedback (e.g., ``Good job!'') or negative feedback. The negative feedback included a slightly frustrating time-out penalty during which participants were forced to wait idly for a duration that was proportional to how sub-optimal their planning operation had been.

\citet{CognitiveTutorsPNAS} found that participants were able to learn to use the automatically discovered strategies, remember them, and use them in novel environments with a similar structure. These findings suggest that automatic strategy discovery can be used to improve human decision-making if the discovered strategies are well-adapted to the situations where people might use them. Additionally, \citet{CognitiveTutorsPNAS} also found that video demonstrations of click sequences performed by the optimal strategy is an equally effective teaching method as providing immediate feedback. Here, we build on these findings to develop cognitive tutors that teach automatically discovered strategies by demonstrating them to people.

\section{Discovering hierarchical planning strategies}
% \js{We have already discussed this part in the introduction. I will comment it out as it is potentially unnecessary.}
% % Humans have the ability to plan far ahead in time, while AI agents 
% % have only been successful to plan for a few steps ahead in time.
% % are still much less capable in doing so.
% % It is important to understand the factors leading to AI agent's poor performance in long horizon planning.
% There are several factors that lead to AI agents' poor performance in long horizon planning when compared to humans. Firstly, as the planning horizon increases, the number of possible action sequences increase exponentially, which makes exhaustive environment exploration computationally intractable. Secondly, the number of possible states in real-world problems are enormous. With the current planning algorithms, the time required to sufficiently explore possible sequences explodes exponentially with \red{number of sequences and states} and becomes computationally infeasible to plan effectively.
% %\js{Another one is that AI is not able to generalize well between domains, but I am not sure whether we should mention that since we are not solving that problem.}

All previous strategy discovery methods evaluate and compare the utilities of all possible computations in each step. As such, these algorithms have to explore the entire metalevel MDP's state space which grows exponentially with the number of nodes.\footnote{The results presented in the paper have upto  $\mathbf{5^{90}}$ possible belief states.}   As a consequence, these methods do not scale well to problems with large state spaces and long planning horizons. This is especially true of the Bayesian Metalevel Policy Search algorithm (BMPS;  \citep{callaway2017learning}) whose run time is exponential in the number of nodes of the planning problem.
% As discussed in previous section,
% \js{Brief introduction to the original BMPS algorithm here or below?}
%\js{It is not evident that we are talking about non-hierarchical methods in the previous sentences. Thus, the transition (contrast) to the part that follows seems unnatural.}
In contrast to the exhaustive enumeration of all possible planning operations performed by those methods, people
would not even consider making detailed low-level motor plans for navigating to a specific distant location (e.g., Terminal C of San Juan Airport) until they arrive at a high-level plan that leads them to or through that location \citep{Tomov2020}. Here, we build on insights about human planning to develop a more scalable method for discovering efficient planning strategies. 
%decompose the problem of maximizing their returns 

\subsection{Hierarchical Problem Decomposition}

\begin{figure}[ht]
  \centering
  \includegraphics[width=0.85\textwidth]{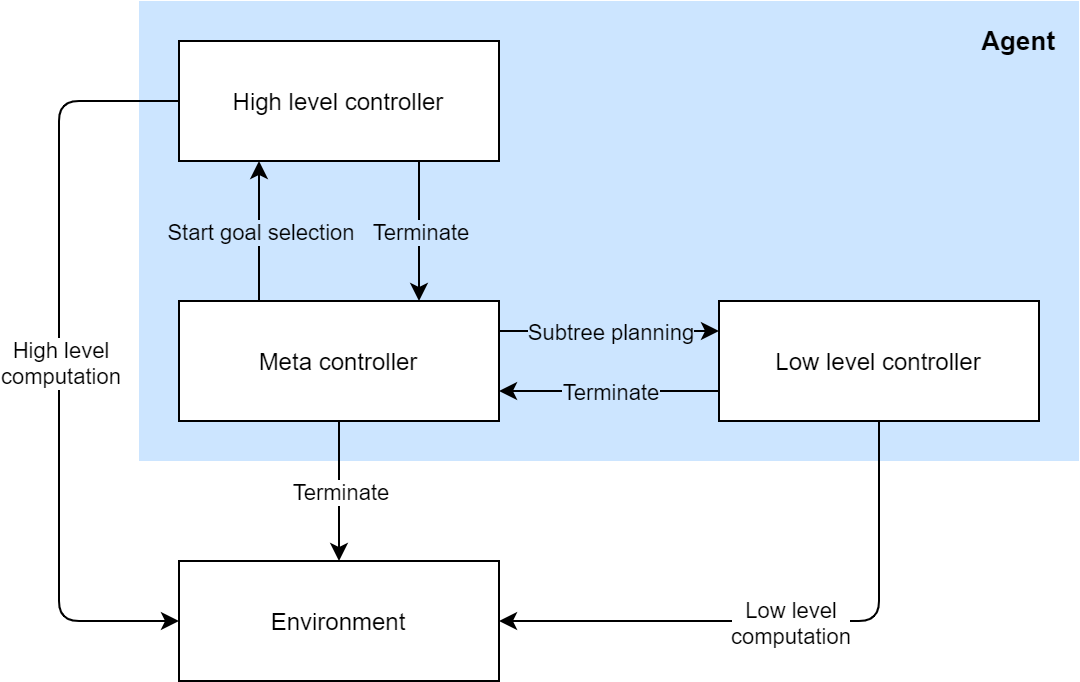}
  \caption{Flow diagram of the hierarchically discovered planning strategy. The high level controller decides on goal computations. The low level controller decides the computations to be performed within a goal.}
  \label{fig:meta}
\end{figure}

To efficiently plan over long horizons, people \citep{Botvinick2008,carver2001self,Tomov2020} and hierarchical planning algorithms \citep{Kaelbling2010,Sacerdoti1974,Marthi2007,Wolfe2010}
decompose the problem into first setting goals and then planning how to achieve them. This two-stage process breaks large planning problems down to smaller problems that are easier to solve. 
To discover hierarchical planning strategies automatically, our proposed strategy discovery algorithm 
% therefore 
decomposes the problem of discovering planning strategies into the sub-problems of discovering a strategy for selecting a goal and discovering a strategy for planning the path to the chosen goal. A pictorial representation is given in Figure~\ref{fig:meta}.

% \begin{figure}[h!]
%   \centering
%   \includegraphics[height=0.25\textwidth]{figs/flow_v4.pdf}  % Two-line flow
% %   \includegraphics[width=\textwidth]{figs/flow_v2s.pdf}  % One-line flow
% %   \includegraphics[width=\textwidth]{figs/flow_v3.pdf}  % One-line flow
%   \caption{Flow diagram of the hierarchically discovered planning strategy. The green boxes refer to portions of the algorithm directly controlled by the planning strategy.}
%   \label{fig:flow}
% \end{figure}
%During the goal setting phase, the algorithm first discovers the optimal strategy to select goal nodes. After setting a goal, the algorithm then discovers the optimal strategy achieve the goal while maximizing the cumulative reward accumulated. 
%By imposing the sequential structure of goal setting and goal achievement,
%This enables our method to discover optimal strategies for large and complex environments much faster than non-hierarchical discovery methods. We now formalize the two sub-problems in turn. %By mimicking human planning behaviour, our solution is an attempt to bring artificial intelligence to human level intelligence. 
%\subsection{Hierarchical decomposition of the algorithm discovery problem}

Formally, this is accomplished by decomposing the metalevel MDP defining the strategy discovery problem into two metalevel MDPs with smaller state and action spaces. Constructing metalevel MDPs for goal-setting and path planning is easy when there is a small set of candidate goals. Such candidate goals can often be identified based on prior knowledge or the structure of the domain \citep{Schapiro2013,solway2014optimal}. A low level controller solves the goal-achievement MDP whereas the high level controller solves the goal-setting MDP. When the controller is in control, a computation is selected from the corresponding metalevel MDP and performed. The meta controller looks at the expected reward of the current goal with the expected reward of the next best goal and  decides when control from the high level controller should be switched to the low level controller. Hence, when the low level controller discovers that the current goal is not as valuable as expected, the meta controller allows for goal switching

The metalevel MDP model of the sub-problem of goal selection (Section~\ref{sec:GoalSettingMDP}) only includes computations for estimating the values of a small set of candidate goal states ($V(g_1),\cdots,V(g_M)$). This means that goals are chosen without considering how costly it would be to achieve them. This makes sense when all goals are known to be achievable and the differences between the values of alternative goal states are substantially larger than the differences between the costs of reaching them. 
% \js{I think the part in the previous sentences needs rephrasing...} 
This is arguably true for many challenges people face in real life. For instance, when a high school student plans one's career, the difference between the long-term values of studying computer science versus becoming a janitor is likely much larger than the difference between the costs of achieving either goal. This is to be expected when the time it will take to achieve the goals is short relative to a person's lifetime.
% In cases where the goals may not be achievable, the meta controller allows the algorithm to retrospect and switch goals, if necessary, to a more realistic goal.

The goal-achievement MDP (Section~\ref{sec:GoalAchievementMDP}) only includes computations that update the estimated costs of alternative paths to the chosen goal by determining the costs or rewards of state-action pairs $r(b,c)$ that lie on those paths. This selection of computations within a selected goal leads to a possible issue of ignoring some computations that can be irrelevant in the goal achievement MDP but be highly valuable when considering the complete problem. One such example is when considering computations which reveal the value of nodes lying on an unavoidable path to the selected goal. This problem gets further accentuated if such a node has a possibility of having a highly positive or negative reward. To rectify this problem, a meta controller has been introduced to facilitate goal switching. 
% To be able to effectively switch goals, it is therefore necessary to incentivize the low level controller to perform certain computations even though they might have no direct value in the low level MDP. 
A real world example of the necessity to switch goals after discovering an unlikely highly negative event could be, for example, to switch from investing in the stock market to investing in real estate after discovering a likely stock market crash.

Decomposing the strategy discovery problem into these two components reduces the number of possible computations that the metareasoning method has to choose between from $M \cdot N$ to $M + N$, where $M$ is the number of possible final destinations (goals) and $N$ is the number of steps to the chosen goal (see Appendix~\ref{sec:time}.  Perhaps the most-promising metareasoning method for automatic strategy discovery is the Bayesian Metalevel Policy Search algorithm (BMPS; \citep{callaway2017learning, KemturJain2020}). To solve the two types of metalevel MDPs introduced below more effectively, we also introduce an improvement of the BMPS algorithm in Section~\ref{sec:BMPS}.

\subsubsection{Goal-setting metalevel MDP} \label{sec:GoalSettingMDP}
The optimal strategy for setting the goal can be formalized as the solution to the metalevel MDP
% $M_{\mathrm{meta, high}} = (\mathcal{B}_{g}, \mathcal{C}^{\mathrm{H}}, \mathcal{T},  r_{\mathrm{meta}}^{\mathrm{H}})$,
$M^{\mathrm{H}} = (\mathcal{B}^{\mathrm{H}}, \mathcal{C}^{\mathrm{H}}, T^{\mathrm{H}},  R^{\mathrm{H}})$,
where the belief state $b^{\mathrm{H}}(g) \in \mathcal{B}^{\mathrm{H}}$ denotes the expected cumulative reward that the agent can attain starting from the goal state $g \in \mathcal{G}$. The high level computations are $\mathcal{C}^{\mathrm{H}} = \{\xi_1, ..., \xi_M, \perp^{\mathrm{H}} \}$, where $\xi_{g}$ reveals the value $V(g)$ of the goal node $g$. $\perp^{\mathrm{H}}$ terminates the high-level planning leading to the agent to select the goal with the highest value according to its current belief state.
The reward function is
% $r_{\mathrm{meta}}^{\mathrm{H}}(b_g, c) = - \lambda^{\mathrm{H}}$
$R^{\mathrm{H}}(b^{\mathrm{H}}, c^{\mathrm{H}}) = - \lambda^{\mathrm{H}}$
for $c^{\mathrm{H}} \in \{\xi_{1}, ... \xi_{M}\}$ and 
% \begin{equation}
% \label{eqn:term-high}
    % r_{\mathrm{meta}}^{\mathrm{H}}(b_{g}, \perp^{\mathrm{H}}) = \max_{k \in \mathcal{G}}\mathbb{E}[b_{g}(k)],
    $
    R^{\mathrm{H}}(b^{\mathrm{H}}, \perp^{\mathrm{H}}) = \max_{k \in \mathcal{G}}\mathbb{E}[b^{\mathrm{H}}(k)]
    $.
% \end{equation}

\subsubsection{Goal-achievement metalevel MDP} \label{sec:GoalAchievementMDP}
Having set a goal to pursue,
% $g \in \mathcal{G}$,
the agent has to find the optimal planning strategy to achieve the goal. This planning strategy is formalized as the solution to the metalevel MDP $M^{\mathrm{L}} = (\mathcal{B}^{\mathrm{L}}, \mathcal{C}^{\mathrm{L}}, T^{\mathrm{L}},  R^{\mathrm{L}})$, where the belief state $b \in \mathcal{B}^{\mathrm{L}}$ denotes the expected reward for each node. The agent can only perform a subset of meta-actions $\mathcal{C}_{g,\mathrm{L}} = \{ c_{g,1}, ..., c_{g,N}, \perp^{\mathrm{L}}\}$, where $c_{g,n}$ reveals the reward at node $n$ in the goal set $h_{g} \in \mathcal{H}$.
A goal set $h_{g} \in \mathcal{H}$ refers to all nodes, including the goal node, which lie on all paths leading to goal $g \in \mathcal{G}$.
Furthermore, $\perp^{\mathrm{L}}$ terminates planning and leads to the agent to select the path with the highest expected sum of rewards according to the current belief state. The reward function is
% $r_{\mathrm{meta}}^{\mathrm{L}}(b, c_{g}) = - \lambda^{\mathrm{L}}$
$R^{\mathrm{L}}(b, c_{g}) = - \lambda^{\mathrm{L}}$
for $c_{g} \in \{ c_{g,1}, ..., c_{g,N}\}$ and 
% \begin{equation}
% \label{eqn:term-low}
    % r_{\mathrm{meta}}^{\mathrm{L}}(b, \perp^{\mathrm{L}}) = \max_{\textbf{t} \in \mathcal{T}} \sum_{k \in \textbf{t}} \mathbb{E}[b_{k}],
    $
    R^{\mathrm{L}}(b, \perp^{\mathrm{L}}) = \max_{p \in \mathcal{P}} \sum_{n \in p} \mathbb{E}[b_{n}]
    $,
% \end{equation}
where $\mathcal{P}$ is the set of all paths, and $b_{n}$ is the belief of the reward for node $n$.

% \subsection{Improvements made to the Bayesian Metalevel Policy Search algorithm} \label{sec:BMPS}
\subsection{Hierarchical Bayesian Metalevel Policy Search} \label{sec:BMPS}

Having introduced the hierarchical problem decomposition, we now present how this decomposition can be leveraged to make BMPS and other automatic strategy discovery methods more scalable. 
%Building on the work by \cite{callaway2017learning}, we extend the BMPS algorithm by introducing hierarchical structure as discussed above. 
BMPS approximates the value of computation (VOC) according to Equation~\ref{eqn:bmps}. We propose to utilize BMPS to solve the goal selection metalevel MDP and the goal achievement metalevel MDP separately. The metacontroller then decides when the policy discovered should run. A detailed analysis of the computational time is presented in the Appendix~\ref{sec:time}.

\paragraph{High level policy search:}

The VOC for the high level policy is approximated using three features: (1) the myopic utility for performing a goal state
% computation 
evaluation ($\mathrm{VOI_{1}^{H}}$), (2) the value of perfect information about all goals ($\mathrm{VPI^{H}}$), and (3) the cost of the respective computation ($\mathrm{cost^{H}}$).
\begin{equation}
\label{eqn:voc-high}
    % \mathrm{\widehat{VOC}}^{\mathrm{H}}(c^{\mathrm{H}},b, b_{g}; \mathbf{w^H}) = w_{1}^{H} \cdot \mathrm{VOI_{1}^{H}}(c^{\mathrm{H}},b_{g}) +  w_{2}^{H} \cdot \mathrm{VPI^{H}}(b_{g}) - w_{3}^{H} \cdot \mathrm{cost^{H}}(c^{\mathrm{H}}),
    \mathrm{\widehat{VOC}}^{\mathrm{H}}(c^{\mathrm{H}}, b^{\mathrm{H}}; \mathbf{w}^{\mathrm{H}}) = w_{1}^{\mathrm{H}} \cdot \mathrm{VOI_{1}^{H}}(c^{\mathrm{H}},b^{\mathrm{H}}) +  w_{2}^{\mathrm{H}} \cdot \mathrm{VPI^{H}}(b^{\mathrm{H}}) - w_{3}^{\mathrm{H}} \cdot \mathrm{cost^{H}}(c^{\mathrm{H}}),
\end{equation}
where $w_{1}^{\mathrm{H}}, w_{2}^{\mathrm{H}}$ are constrained to a probability simplex set,
% where $w_{1}^{H}$, $w_{2}^{H} \in \mathbb{R}_{\ge 0}$, $w_{1}^{H} + w_{2}^{H} = 1$
$w_{3}^{\mathrm{H}} \in \mathbb{R}_{[1, M]}$, and $M$ is the number of goals. Additionally, the cost $\mathrm{cost^{H}}(c^{\mathrm{H}})$ is defined as
% \subsubsection{Low level policy search}
% \js{I added statements about the constraints in Section~\ref{sec:MouselabMDP}. Do we have to write that here too?}
\begin{equation}
\label{eqn:high-cost}
    \mathrm{cost^{H}}(c^{\mathrm{H}})=\begin{cases}
    \lambda^{\mathrm{H}}, & \text{if $c^{\mathrm{H}} \in \{\xi_{1},..., \xi_{M} \}$}.\\
    0, & \text{if $c^{\mathrm{H}} = \perp^{\mathrm{H}}$}.
  \end{cases}
\end{equation}

\paragraph{Low level policy search:}

In a similar manner as for the high-level policy,
% BMPS approximates
the value of computation for the low-level policy 
% within a goal set
is approximated by using a mixture of VOI features and the anticipated
% modified
cost of the current computation and future computations, that is:
\begin{align}
% \begin{gathered}
    \label{eqn:voc-low}
    \mathrm{\widehat{VOC}}^{\mathrm{L}}(c,b, g; \mathbf{w}^{\mathrm{L}})
    & = w_{1}^{\mathrm{L}} \cdot \mathrm{VOI_{1}^{L}}(c,b,g) +  w_{2}^{\mathrm{L}} \cdot \mathrm{VPI^{L}}(b,g)  \\
    & + w_{3}^{\mathrm{L}} \cdot \mathrm{VPI_{sub}^{L}}(c,b,g) - w_{4}^{\mathrm{L}} \cdot \mathrm{cost^{L}}(c,g, \mathbf{w}^{\mathrm{L}})  \nonumber
% \end{gathered}
\end{align}

where $w_{i}^{\mathrm{L}}$ ($i = 1, 2, 3$) are constrained to a probability simplex set,
% where $w_{i}^{L} \in \mathbb{R}_{\ge 0}$ ($i = 1, 2, 3$), $\sum_{i=1}^{3} w_{i}^{L} = 1$,
% with the constraints that $w_{1}^{L}, w_{2}^{L}, w_{3}^{L} \in [0,1]$, $w_{1}^{L} + w_{2}^{L} + w_{3}^{L} = 1$,
$w_{4}^{\mathrm{L}} \in \mathbb{R}_{[1, |h_{g}|]}$, and $|h_{g}|$ is the number of nodes in goal set $h_{g}$. The weight values for both levels are optimised in $100$ iterations with Bayesian Optimization \citep{mockus2012bayesian} using the GPyOpt library \citep{gpyopt2016}.

The cost feature of the original BMPS algorithm introduced by \citet{callaway2017learning} only considered the cost of a single computation whereas its VOI features consider the benefits of performing a sequence of computations. As a consequence, policies learned with the original version of BMPS are biased towards inspecting nodes that many paths converge on, even when the values of those nodes are irrelevant. To rectify this problem,  we redefine the cost feature so that it considers the costs of all
% of the
computations assumed by the VOI features. Concretely, to compute the low-level policy, we define the cost feature of BMPS as the weighted average of the costs of generating the information assumed by the VOI features $\mathcal{F}=\lbrace \mathrm{VOI_{1}^{L}}, \mathrm{VPI^{L}}, \mathrm{VPI_{sub}^{L}}\rbrace$, that is
\begin{equation}
\label{eqn:cost}
    \mathrm{cost^{L}}(c,g, \mathbf{w}^{\mathrm{L}}) = \sum_{f \in \mathcal{F}} w_{f}^{\mathrm{L}} \cdot \sum_{n}^{|h_{g}|} \mathbb{I}(c,f,n) \cdot \mathrm{cost}(c)
\end{equation}
where $\mathbb{I}(c,f,n)$ returns $1$ if node $n$ is relevant when computing feature $f$ for computation $c$ and $0$ otherwise.

% The index $i$ iterates from $1$ to $3$ corresponding to features $\mathrm{VOI_{1}^{L}}$, $\mathrm{VPI^{L}}$, and $\mathrm{VPI_{sub}^{L}}$ respectively.
% \red{This modification reduces the dominance of $\mathrm{VPI_{sub}^{L}}$ when selecting initial computations.}
% in the beginning. 
%This slight increment in penalization to the VOC reduces the dominance of $\mathrm{VPI_{sub}^{L}}$ when selecting initial computations. It finds an optimal low level policy
% search
%invariant to the goal node value, since the optimal low level policy should not make computations related to the goal node. Intuitively, this is to be expected, since the goal node is a child node for all potential paths within a goal set and visiting the goal node is unavoidable irrespective of its concealed value. 

%Since when the low level controller is running, the expected reward of the unselected goals will not change, no additional computational resources are utilized.

In the remainder of this article, we refer to the resulting strategy discovery algorithm as \textit{hierarchical BMPS} and refer to the original version of BMPS as \textit{non-hierarchical BMPS}.

\subsection{Tree contraction method for faster BMPS feature computation} \label{sec:tree_contraction}

To further increase the scalability of BMPS, we make an additional improvement to how it computes the features used to approximate the value of computation \citep{callaway2017learning}. 
Specifically, we aim to improve the computational efficiency by combining nodes in the meta MDP according to a set of predefined conditions, ultimately reducing the complexity of the necessary computations. The node combination is performed by merging two nodes into a single new node with a probability distribution that represents their combined reward value.  

The algorithm consists of three different operations that combine node distributions.
% Which operation to apply is determined by 
A list of conditions determine an operation to apply and the algorithm stops when the distributions of all nodes within the MDP are collapsed to a single root node.
\begin{itemize}
    \item \textbf{Add}: Combines the distribution of two consecutive nodes\footnote{Two nodes are consecutive if they are in a direct parent-child relation.} by adding their distributions. This operation can be applied to two consecutive nodes in the tree as long as the parent node does not have other child nodes and the child node does not have other parents. 
    \item \textbf{Maximise}: Combines two parallel nodes by taking the maximum value for each combination of values of nodes can take, combining the nodes distributions while taking into account that the optimal path will always lead through the higher node of the two. This operation can be applied to two nodes that have a single identical parent and child node.
    \item \textbf{Split}: Splits a child node into two separate nodes by duplicating that node. The whole tree is then duplicated as many times as the node has possible values, fixing the node's distribution to each possibility. The duplicated trees are then individually reduced to single root nodes and the individual root nodes are combined to a single tree by pairwise application of the maximise operation. This operation can be applied to nodes that have multiple parent nodes where each of the individual nodes after splitting is only connected to one its parent nodes.
\end{itemize}

The split operation is the most computationally expensive operation and is therefore only applied when the add and maximise operations are insufficient to reduce the tree to a single node. Specifically, this happens when a node that needs to be reduced by the multiply operation has an additional parent or child node. Since the structure of the environment stays identical while the rewards and discovered states vary, we precompute the necessary operations to reduce the tree and then apply the reduction individually for each problem instance.

Our adjustment is purely algorithmic and it does not change the value of computation. Therefore, it does not impair the performance of the discovered strategies.
An additional effect of the tree contraction method is that it extends the types of environments solvable by BMPS. Previously, BMPS was only able to handle environments with a branching tree structure: nodes can have multiple children but never multiple parents. Our new formulation allows us to compute the BMPS features for tree structures in which nodes have multiple parent nodes as well. This is possible through the application of the maximise operation, which allows to combine multiple parent nodes into a single node, making them solvable through the value of computation calculation. The range of solvable environments is therefore extended from trees to directed acyclic graphs. This extension is especially relevant for environments containing goal nodes since it is often the case that multiple intermediate nodes converge to the same goal node.

\section{Evaluating the performance, scalability, and robustness of our method for discovering hierarchical planning strategies}\label{sec:benchmarks}

To evaluate our method for discovering hierarchical planning strategies, we benchmarked its performance, scalability, and robustness using the two types of environments illustrated in Figure~\ref{fig:env} and Figure~\ref{fig:high_risk}. The first type of environments conforms to the structure that motivated our hierarchical problem decomposition (i.e., the variability of rewards increases from each step to the next) and the second type does not. In the second type of environment, we introduced a high-risk node on the path to each goal (see Figure~\ref{fig:high_risk}). This violates the assumption that motivated the hierarchical decomposition and makes goal switching essential for good performance. 

For each environment, we apply the criterion defined in Equation~\ref{eq:RR} (see Sections~\ref{sec:RR} and \ref{sec:MouselabMDP}) to evaluate the degree to which the resulting strategies are resource rational against the resource rationality of human planning, existing planning algorithms, and the strategies discovered by state-of-the-art strategy methods. Furthermore, we show the our method is substantially more scalable than previous methods.

To be able to compare the performance of the automatically discovered planning strategies to the performance of people, we conduct experiments on Amazon Mechanical Turk \citep{litman2017turkprime}. In these experiments, we measure human performance in Flight Planning tasks that are analogous to the environments we use to evaluate our method (see Figure~\ref{fig:FlightPlanningTask}). For the first type of environments, we recruited $78$ participants for each of the four environments (average age $34.71$ years, range: 19–70 years; $46$ female). Participants were paid \$$2.00$ plus a performance-dependent bonus (average bonus \$$1.52$). The average duration of the experiment was $25.1$ min. 
For the second type of environments, we recruited $48$ participants (average age $36.98$ years, range: 19–70 years; $25$ female). Participants were paid \$$1.75$ plus a performance-dependent bonus (average bonus \$$0.34$). The average duration of the experiment was $14.86$ min.
Following instructions that informed the participants about the range of possible reward values, participants were given the opportunity to familiarize themselves with the task in $5$ practice trials of the Flight Planning task. After this, participants were evaluated on $15$ test trials of the Flight Planning task for the first type of environments and $5$ test trials for the second type of environments. To ensure high data quality, we applied the same pre-determined exclusion criterion throughout all presented experiments. We excluded participants who did not make a single click on more than half of the test trials because not clicking is highly indicative to a participant not engaging and speeding through the task. In the first environment type we excluded $3$ participants and in the second environment type we excluded $22$ participants.

\subsection{Evaluation in environments that conform to the assumed structure} \label{sec:hierarchical_eval}

We first evaluate the performance and scalability of our method in environments whose structure conforms to the assumptions that motivated the hierarchical problem decomposition. To do so, we compare the performance of the discovered strategies against the performance of existing planning algorithms, the strategies discovered by previous strategy discover methods, and human performance in four increasingly challenging environments of this type with 2-5 candidate goals. The reward of each node is sampled from a normal distribution with mean $0$. The variance of rewards available at non-goal nodes was $5$ for nodes reachable within a single step (level 1) and doubled from each level to the next. The variance of the distribution from which the reward associated with the goal node was sampled starts from $100$ and increases by $20$ for every additional goal node. The environment was partitioned into one sub-graph per goal. Each of those sub-graphs contains 17 intermediate nodes, forming 10 possible paths that reach the goal state in maximum 5 steps (see Figure~\ref{fig:env}). The cost of planning is $1$ point per click ($\lambda=1$).

To estimate an upper bound on the performance of existing planning algorithms on our benchmark problems, we selected Backward Search and Bidirectional Search \citep{russell2002artificial} because -- unlike most planning algorithms -- they start by considering potential final destinations, which is optimal for planning in our benchmark problems. 
These search algorithms terminate when they find a path whose expected return exceeds a threshold, called its aspiration value. The aspiration was selected using Bayesian Optimization \citep{mockus2012bayesian} to get the best possible performance from the selected planning algorithm.
% as described in Section~\ref{sec:benchmarks}.
We also evaluated the performance of a random-search algorithm, which chooses computations uniformly at random from the set of metalevel operations that have not been performed yet.

In addition to those planning algorithms, our baselines also include three state-of-the-art methods for automatic strategy discovery: the greedy myopic VOC strategy discovery algorithm \citep{lin2015metareasoning}, which approximates the VOC by its myopic utility ($\text{VOI}_1$), BMPS \citep{callaway2017learning}, and the Adaptive Metareasoning Policy Search algorithm (AMPS) \citep{svegliato2018adaptive} which uses approximate metareasoning to decide when to terminate planning. Our implementation of the AMPS algorithm
% adaptive metareasoning policy search algorithm
uses a deep Q-network \citep{mnih2013playing} to estimate the difference between values to stop planning and continue planning, respectively. It learns this estimate based on
the expected termination reward of the best path. The planning operations are selected by maximizing the myopic value of information ($\text{VOI}_1$). 
% We compared the performance of these algorithms with our strategy discovery algorithm which searches for policies in a hierarchical manner.
When applying hierarchical BMPS to this environment, we disabled the goal switching component of the metacontroller since the cumulative variance of the intermediate nodes was less than the variance of the goal nodes, rendering goal-switching unnecessary.
To illustrate the versatility of our hierarchical problem decomposition, we also applied it to the greedy myopic VOC strategy discovery algorithm. 

% This hierarchical decomposition, drastically reduces the number of calculations needed for computation selection and is far more scalable than the present planning algorithms. 
% Apart from hierarchical BMPS, we also evaluated the performance of the hierarchical policy search algorithm utilising myopic VOC.

\subsubsection{Performance of the automatically discovered strategies}
\label{sec:results}

\begin{figure}[ht]
  \centering
  \includegraphics[width=\textwidth]{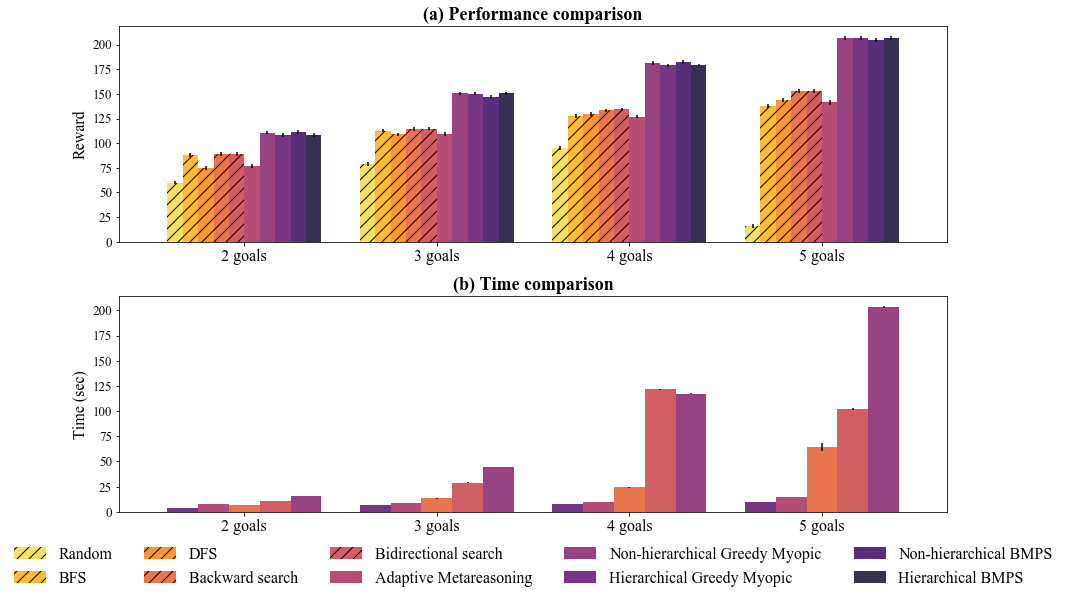}
  \caption{(a) Net returns of existing planning algorithms (striped bars) versus planning strategies discovered by various strategy discovery methods (bars without stripes). (b) Comparison of the mean time for various strategy discovery methods (in seconds).}
  \label{fig:eval-merged}
\end{figure}

Table~\ref{tab:reward} and Figure~\ref{fig:eval-merged}a compare the performance of the strategies discovered by hierarchical BMPS and the hierarchical greedy myopic VOC method against the performance of the strategies discovered by the two state-of-the-art methods, two standard planning algorithms, and human performance on the benchmark problems described above (Section~\ref{sec:hierarchical_eval}).
% showed that the automatically discovered strategies outperform standard planning strategies. 
These results show that the strategies discovered by our new hierarchical strategy discovery methods outperform extant planning algorithms and the strategies discovered by the AMPS algorithm across all of our benchmark problems ($p<.01$ for all pairwise Wilcoxon rank-sum tests). Critically, 
% while 
imposing hierarchical constraints on the strategy search of BMPS and the greedy myopic VOC method had no negative effect on the performance of the resulting strategies ($p > .770$ for all pairwise Wilcoxon rank-sum tests). Additionally, when human participants were tested on $15$ environments, they performed much worse than than the performance of the strategy discovered by our hierarchical method regardless of the number of goals ($p < 0.02$ for all pairwise Wilcoxon rank-sum tests).

\begin{table}[ht]
  \centering
  \begin{tabular}{clcccc}
    \toprule
    Type & Name                                 & 2 Goals           & 3 Goals           & 4 Goals           & 5 Goals\\
    \midrule
    $S$ & \textbf{Hierarchical BMPS}                     & 108.79            & \textbf{150.63}   & 178.98            & 206.45 \\
    $S$ & \textbf{Non-hierarchical BMPS}                 & \textbf{111.53}   & 148.01            & \textbf{182.38}   & 204.37 \\
    $S$ & \textbf{Hierarchical greedy myopic VOC}        & 108.48            & 150.13            & 178.81            & \textbf{206.57} \\
    $S$ & \textbf{Non-hierarchical greedy myopic VOC}    & 107.98            & 150.41            & 180.35            & 205.40 \\
    $S$ & Adaptive Metareasoning Policy Search  & 77.08             & 109.39            & 127.01            & 141.34 \\
    \midrule
     $P$ & Depth-first Search                  & 74.99             & 109.13           & 129.66           & 143.45 \\
     $P$ & Breadth-first Search                  & 87.62             & 112.83           & 127.68           & 137.40 \\
    $P$ & Bidirectional Search                  & 88.59             & 115.07            & 134.08            & 154.24 \\
    $P$ & Backward Search                       & 87.85             & 114.29            & 134.43            & 156.56 \\
    $P$ & Random Policy                         & 52.73             & 80.05             & 89.31             & 101.15 \\
    \midrule
   & Human Baseline                             & 45.42             & 88.06             & 39.32             & 124.89\\
    \bottomrule
  \end{tabular}
  \caption{Net returns of various strategy discovery methods ($S$) and existing planning algorithms ($P$). The best algorithms and the best net returns for each environment setting (column) are formatted in \textbf{bold}. The four best methods performed significantly better than the other methods but the differences between them are not statistically significant. %$S$: strategy discovery algorithm; $P$: planning algorithm.
  }
  \label{tab:reward}
\end{table}

% To highlight the difference in performance of discovering algorithms in a hierarchical and non-hierarchical manner, Figures \ref{fig:option-reward} and \ref{fig:train-time} showcase the similarity in mean reward for BMPS with hierarchical and non-hierarchical planning and the huge difference in time required to train for just 10 iterations of Bayesian Optimization.
% although
%and the hierarchical BMPS significantly outperformed the
% adaptive metareasoning
%AMPS algorithm.
%Critically, as shown in Figure~\ref{fig:eval-merged}a and Table~\ref{tab:reward}, this improved scalability is achieved without comprimising the performance of either BMPS or the greedy myopic VOC strategy discovery method.

%In  the 5-goal environment, the hierarchical greedy myopic VOC strategy discovery algorithm has a net return of $206.57 \pm 2.03 $ and the backward search has a net return of $152.95 \pm 2.07$ ($t(198) = 3.6210$, $p < .001;$ effect size $d = 0.514$). 

\begin{figure}[ht]
  \centering
  \includegraphics[height=0.45\textwidth]{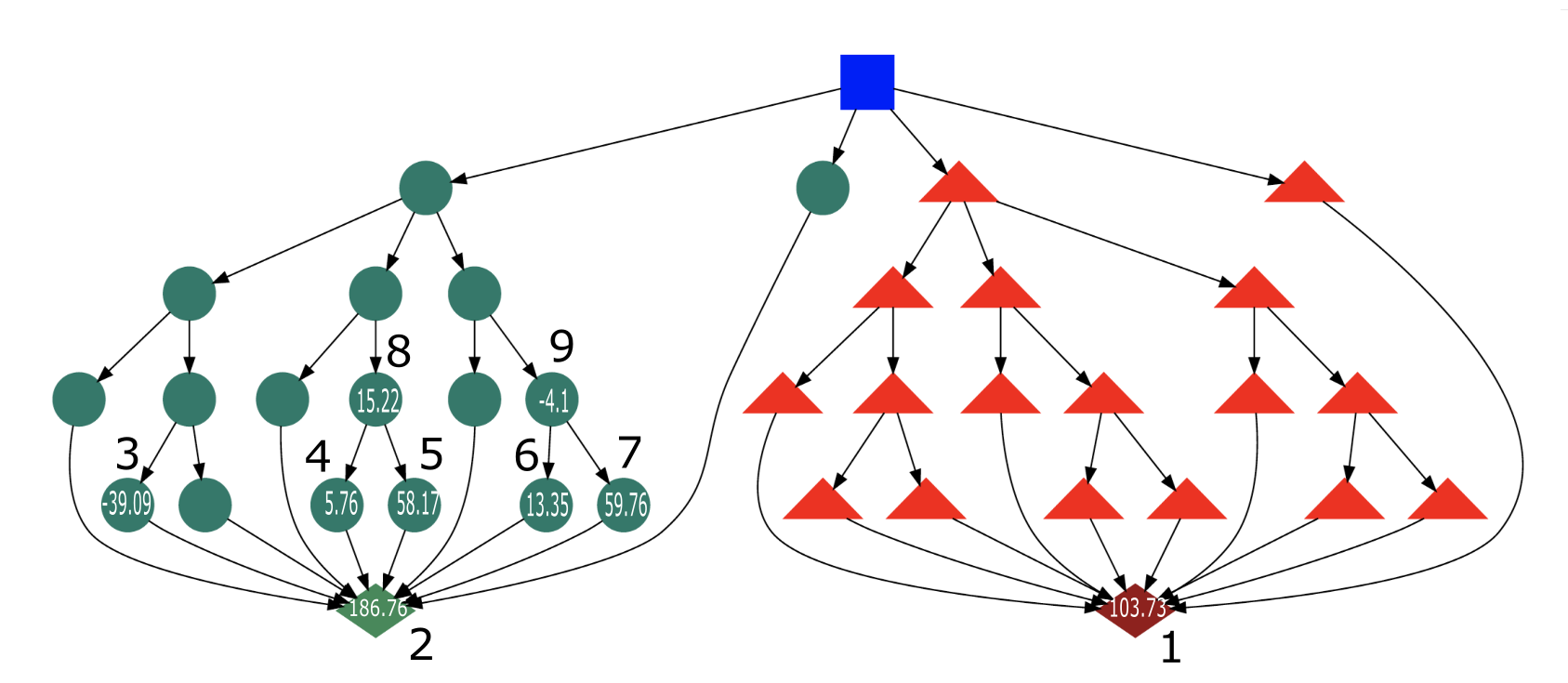}  % Two-line flow
  \caption{Sequence of nodes revealed in particular environment. The numbers above the nodes indicate the sequence in which the nodes were revealed. The numbers in each revealed node indicates its reward.}
  \label{fig:clicks}
\end{figure}

As illustrated in Figure~\ref{fig:clicks}, the planning strategy  our hierarchical BMPS algorithm discovered for this type of environment is qualitatively different from all existing planning algorithms In general, the strategy is as follows:  it first evaluates the goal nodes until it finds a goal node with a sufficiently high reward. Then, it plans backward from the chosen goal to the current state. In evaluating candidate paths from the goal to the current state, it
% discounting
discards each path from further exploration as soon as it encounters a high negative reward on that path. This phenomenon is known as pruning and has previously been observed in human planning (Huys, et al., 2012)\nocite{Huys2012}.\footnote{While this strategy was discovered assuming that the cost of evaluating a potential goal node is the same as the cost of evaluating an intermediate node, we found that the discovered strategy remained the same as we increased the cost of evaluating goal nodes to  $2$, $5$, or $10$.} The non-hierarchical version of BMPS also discovered this type of planning strategy. This suggests that goal-setting with backward planning is the resource-rational strategy for this environment rather than an artifact of our hierarchical problem decomposition. Unlike this type of planning, most extant planning algorithms plan forward and the few planning algorithms that plan backward (e.g., Bidirectional Search and Backward Search) do not preemptively terminate a path exploration. 
%Also, it is seen that the evaluation of goal nodes before intermediate nodes is not  an artifact of the imposed hierarchical structure, which means that with or without the hierarchical structure, the evaluataion of goal nodes are preferred over intermediate nodes for planning. This is justified by the strategy discovered by the non-hierarchical BMPS algorithm which followed a similar pattern.  Rather, they reflect the nature of the optimal planning algorithm for environments where goal values are more variable than immediate and intermediate rewards.  %It is important to note that
%In fact, we found that even though the hierarchical decomposition eliminated a large number of possible planning strategies from the set of possible strategies, it retained the optimal planning algorithm for the studied environment. % the algorithms discovered with the hierarchical decomposition were just as good as those discovered without this simplification.

% , which implies that the optimal planning strategy is implicitly hierarchical in nature.
% The statistical significance in the results are validated by the small interval of the error bars in Figure~\ref{fig:eval-merged}(a) which represent the $95\%$ confidence bound of the mean cumulative reward. 

\subsubsection{Scalability}

Table~\ref{tab:eval-time} and Figure~\ref{fig:eval-merged}b compare the run times of our hierarchical strategy discovery methods against their non-hierarchical counterparts and adaptive metareasoning policy search. This comparison shows that imposing hierarchical structure substantially increased the scalability of BMPS and the greedy myopic VOC method. %is substantially increased by imposing hierarchical constraints, which is evident from the reduced run time and time complexity of both strategy discovery methods (see Figure~\ref{fig:eval-merged}b and Table~\ref{tab:eval-time}).
%This speed-up increased the size of planning problems for which resource-rational planning strategies can be discovered by a factor of X from environments with at most A nodes to environments with up to B nodes. 
The improved run time profile reflects a reduction in the asymptotic upper bound on the algorithms' run times when  hierarchical structure is imposed on the strategy space (see Appendix ~\ref{sec:time}). 
%Additionally, by comparing the largest environment for which the optimal strategy is discovered by the hierarchical and non-hierarchical strategy discovery algorithms, we verify the applicability of the scalable algorithm in real-life problems. 
%The last column of Table~\ref{tab:eval-time} shows that this speed 
As shown in the last column of Table~\ref{tab:eval-time}, this reduction in computational complexity increases the size of environments for which we can discovery resource-rational planning strategies by a factor of 14-15, depending on the required quality of the planning strategy. This makes it possible to automatically discover planning strategies for sequential decision problems with up to $2520$ states. Consequently, our method scales to metalevel MDPs with up to $5^{2520}$ possible belief states whereas the original version of BMPS was limited to problems with only up to $5^{36}$ possible belief states \footnote{The continuous normal distribution is discretized to 4 bins. So including the undiscovered state, each node has 5 possible state conditions}. This shows that our approach increased the scalability of automatic algorithm discovery by a factor of $\mathbf{5^{2484}}$. This is a significant step towards discovering resource-rational planning strategies for the complex planning problems people face in the real world. %se further shown theoretically in the Appendix~\ref{sec:time}.
While the hierarchical greedy myopic VOC method is the most scalable strategy discovery method, the fastest method on our four benchmarks was our hierarchical BMPS algorithm with tree contraction. Comparing the first two rows shows that the tree contraction method described in Section \ref{sec:tree_contraction} significantly contributed to this speed-up (for more details see Appendix~\ref{sec:tree-contraction-speedup}). 
\begin{table}[ht]
  \centering
   \begin{tabular}{lccccc}
    \toprule
    Strategy Discovery Algorithm                & 2 Goals           & 3 Goals       & 4 Goals       & 5 Goals & Max. \#~Nodes\\
    \midrule
    \textbf{Hierarchical BMPS}                           & \textbf{0.21}     & \textbf{0.23} & \textbf{0.24} & \textbf{0.27} & 540\\
    \textbf{with tree contraction} &      & & & & \\
    Hierarchical BMPS                         & 4.18     & 6.45 & 7.45 & 9.30 & 180\\
    Non-hierarchical BMPS                       & 16.09             & 44.81         & 117.46        & 203.18 & 36\\
    \textbf{Hierarchical greedy}  & 7.35  & 8.52          & 10.03         & 14.64 & \textbf{2520}\\
    \textbf{myopic VOC} & & & & & \\
    Non-hierarchical greedy       & 10.56             & 29.02         & 121.53        & 101.97 & 180\\
    myopic VOC &      & & & & \\
    Adaptive Metareasoning         & 6.45              & 13.39         & 24.43         & 64.36 & 36\\
    Policy Search &      & & & & \\
    \bottomrule
  \end{tabular}
  \caption{Average time to evaluate an environment represented in seconds. The last column denotes the size of the largest environment (\#~nodes) for which each method can planning strategies within a time budget of 8~h.}
  \label{tab:eval-time}
\end{table}

\subsection{Robustness to violations of the assumed structure} \label{sec:metacontroller_eval}

\begin{table}[!t]
  \centering
%   \begin{tabular}{lllll}
%     \toprule
%     Ratio ($A:B,C$) & Hierarchical & Non-hierarchical\\
%     \midrule
%     75: 100, 120 & 108.79 & 107.98 \\
%     75: 75, 75 & 89.41 & 86.59 \\
%     75: 50, 60 & 76.84 & 75.51 \\
%     75: 25, 30 & 60.62 & 61.92 \\
%     75: 12, 15 & 53.63 & 59.16 \\
%     \bottomrule
%   \end{tabular}

  \begin{tabular}{ccccccc}
    \toprule
    $\sigma_{\Sigma}$ & $\sigma_{1}$ & $\sigma_{2}$ & Hierarchical & Non-hierarchical & $\Delta$ & $ \% \Delta$  \\
    \midrule
    46.1 & 100 & 120 & 108.79 & 111.53 & 2.74 & 2.46\\
    46.1 & 75 & 75 & 89.41 & 90.62 & 1.21 & 1.34\\
    46.1 & 50 & 60 & 76.84 & 79.52 & 2.68 & 3.37\\
    46.1 & 25 & 30 & 60.62 & 64.95 & 4.33 & 6.67\\
    46.1 & 12 & 15 & 53.63 & 59.63 & 6.00 & 10.06\\
    \bottomrule
  \end{tabular}
  \caption{Comparison in performance of BMPS with and without hierarchical structure on 2-goal environment with various variance ratios. $\sigma_{\Sigma}$: cumulative standard deviation of the longest path to a goal node; $\sigma_{1}$: standard deviation of the first goal node; $\sigma_{2}$: standard deviation of the second goal node; $\Delta$: absolute difference between net returns; $\% \Delta$: relative difference between net returns.}
  \label{tab:var}
\end{table}
% \js{We can remove the first column and say in the text that we fix this value to 75.}

% \begin{figure}
%   \centering
%   \includegraphics[height=0.50\textwidth, width=\textwidth]{figs/T-test.eps}
%   \caption{Performance of different algorithms}
%   \label{fig:reward}
% \end{figure}

To determine the range of planning problems for which our hierarchical decomposition can be used to discover resource-rational planning strategies at scale, we varied the variance structure of the 2-goal environment and compared the performance of BMPS with and without hierarchical structure (see  Table~\ref{tab:var}). 
% The notation, $A: B, C$, used to denote the ratio in Table \ref{tab:var} represents the cumulative variance of the longest intra-goal nodes, the variance of the first goal node and the variance of the second goal node, respectively.
We found that the usefulness of the hierarchical decomposition depends on the variance ratios of the goal values versus path costs. %The results corroborate this and show that 
Concretely, the performance loss of the algorithm utilising the hierarchical structure is below $5\%$ when the variance of goal values is at least as high as the variance of the path costs, but increases to $10\%$ as the variance of goal values drops to only one third of the variance of the path costs.
% within a goal.

\begin{figure}[ht]
  \centering
  \includegraphics[width=\textwidth]{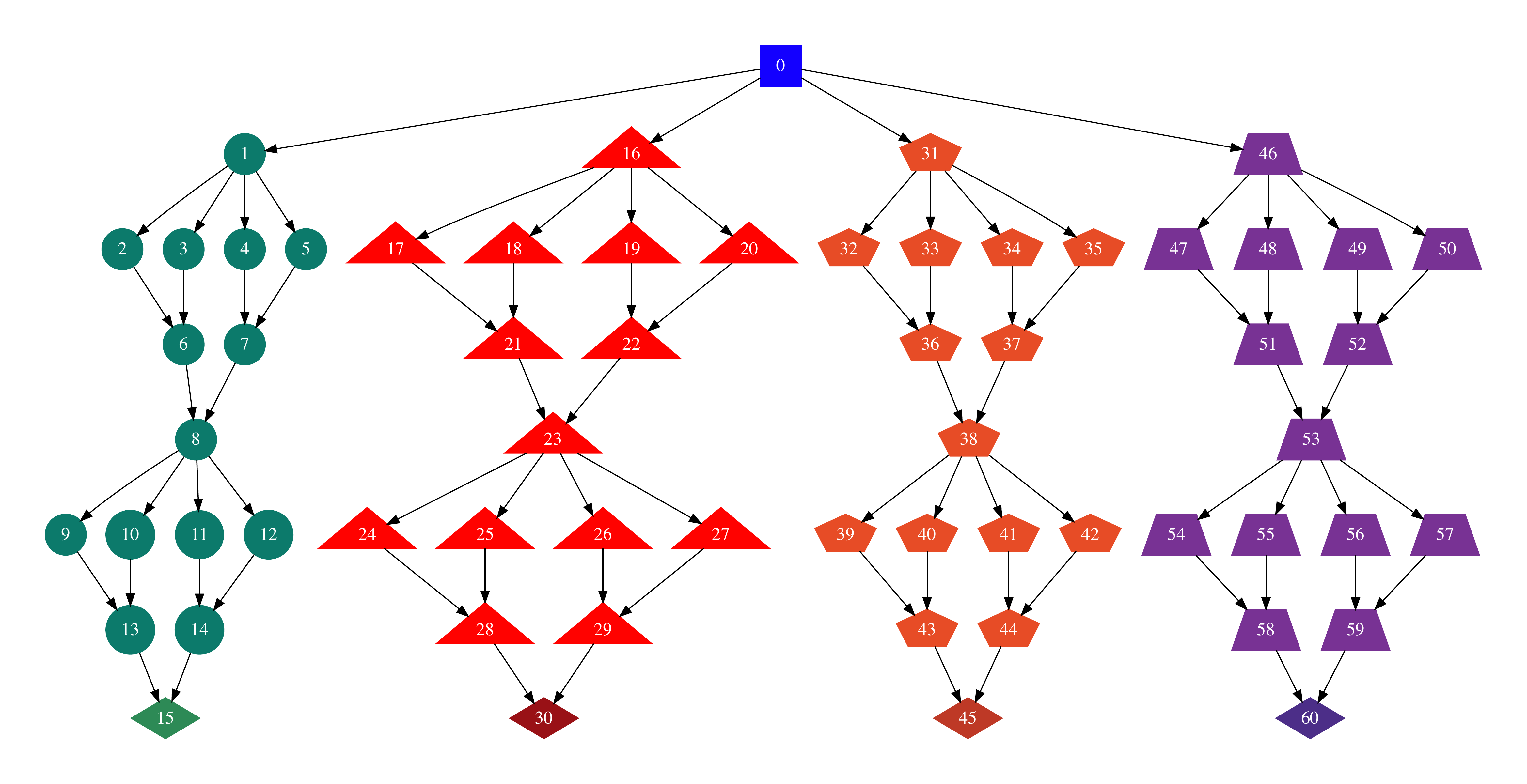}
  \caption{Environment that demonstrates the utility of goal switching. High risk nodes (8, 23, 38 and 53) follow a categorical reward distribution of -1500 with a probability of 0.1 and 0 with a probability of 0.9. Goal nodes (15, 30, 45 and 60) have a categorical equiprobable reward distribution of 0, 25, 75 or 100. The first node in each sub-tree (1, 16, 31 and 46) as well as the root node (0) have a fixed reward of 0. All other intermediate nodes follow a categorical equiprobable distribution of -10, -5, 5, 10.}
  \label{fig:high_risk}
\end{figure}

%METACONTOLLER EVALUATION
To accommodate environments whose structure violates the assumption that more distant rewards are more variable than more proximal ones, the hierarchical strategies discovered by our method can alternate between goal selection and goal planning. 
We now demonstrate the benefits of this goal switching functionality by comparing the performance of our method with versus without goal switching. In particular, we demonstrate that switching goals leads to a better performance if the assumption of increasing variance is violated and does not harm performance when that assumption is met.%, making the goal switching algorithm strictly better than the lessened version without goal switching.
%To demonstrate the effectiveness of goal switching, we compared the performance against our method without goal switching. 
Firstly, we compare the performance of the two algorithms in an environment where switching goals should lead to an improvement in performance. This environment has a total of 60 nodes split into four different goals, each consisting of 15 nodes in the low-level MDP. The difference to previously used environments is that one of the unavoidable intermediate nodes has a 10\% to harbor a large loss of -1500 (see Figure \ref{fig:high_risk}).The cost of computation in this environment is 10 points per click (i.e., $\lambda=10$).The optimal strategy for this environment selects a goal, checks this high-risk node on the path leading to the selected goal and switches to a different goal if it uncovers the large loss. 
% Also compared to vanilla BMPS
We compare the performance of hierarchical BMPS with goal-switching to the performance of hierarchical BMPS without goal-switching, the non-hierarchical BMPS method with tree contraction\footnote{Without our tree contraction method, the original version of BMPS would not have been scalable enough to handle this environment.}, and human performance. The three strategy discovery algorithms were all trained on the same environment following the same training steps. Their performance is noted in Table \ref{table:sim_high_risk}. 

%We also compared the performance of our method to human performance. 
%To gather a human baseline, we conducted a small online experiment in which \lovis{TODO} participants were recruited on Amazon's Mechanical Turk \cite{litman2017turkprime}. Participants were first shown 5 training trials in which they were able to explore the environment and learn a strategy, and then evaluated in 5 test trials. 
Since humans were evaluated on only 5 randomly selected instances of this environments and each environment instance contains some randomness in its reward values, we evaluated the performance of the strategy discovery methods on the same 5 environments. 
% show better performance when switching goals is worth it
All performance scores do not follow a normal distribution as tested with a Shapiro-Wilk test ($p<.001$ for each). The performance between the individual algorithms was compared with a Wilcoxon rank-Sum rest, adjusting the critical alpha value via Bonferroni correction.  
Comparing the score of goal-switching to both our method without goal-switching ($W=14.07, p<.001$) and the original BMPS algorithm (W=11.38, p<.001), shows a significant benefit of goal-switching. Comparing the performance of the original BMPS method to the no goal-switching algorithm, the original BMPS version performs significantly better ($W=18.7, p<.001$).
While the  average human performance was only $-79.92$ (see Table \ref{table:sim_high_risk}), our method achieved a resource-rationality score of $41$ on the same environment instances. A Shapiro-Wilk detected no significant violation of the assumption that participants' average scores are normally distributed ($p.=33$). We therefore compared the average human performance to our method in a one-sample t-test. We found that human participants performed significantly worse than the strategy discovered by our method ($t(25)=-13.06, p<.001$). This suggests that the strategy discovered by our method achieved a superhuman level of computational efficiency.

\begin{table}[ht]
\centering
\begin{tabular}{l|ccc}
\toprule
Algorithm       & N    & Reward & Std    \\ 
\midrule
No goal-switching & 5000 & -80.38      & 446.47 \\ 
Goal-switching    & 5000 & 51.33       & 32.2  \\ 
Non-hierarchical BMPS & 5000 & 39.29 & 40.93\\
\midrule
Human baseline & 26 & -79.92 & 74.06\\
\bottomrule
\end{tabular}
\caption{Mean reward and standard deviation of executing the the hierarchical planning algorithm with and without goal-switching, as well as the original BMPS algorithm, over 5000 random instances of the high-risk environment with four goal states. A human baseline is gathered in an online experiment with a lower number of samples.}
\label{table:sim_high_risk}
\end{table}

To show that enabling our algorithm's capacity for goal-switching has no negative effect on its performance even when the assumption of the hierarchical decomposition is met, we perform a second comparison on the two-goal environment with increasing variance as in Section \ref{sec:results}. Since in this environment the rewards are most variable at the goal nodes, 
% this environment has a reward structure where the variance is the highest in the goal nodes, 
switching goals should usually be unnecessary. Therefore, due to the environment structure, we do not expect the goal-switching strategy to perform better than the purely hierarchical strategy. By comparing the performance in this environment we observe that both versions of the algorithm perform similarly well \ref{table:sim_low_risk}. A Wilcoxon rank-Sum rest ($W=0.03, p=.98$) shows no significant difference between the two. This demonstrates that the addition of goal switching to the algorithm does not impair performance, even when goal switching is not beneficial.

\begin{table}[ht]
\centering
\begin{tabular}{l|ccc}
\toprule
Algorithm       & N    & Reward & Std    \\ 
\midrule
No goal-switching & 5000 &   108.84    & 95.37 \\ 
Goal-switching    & 5000 &    108.78   & 95.37  \\ 
\bottomrule
\end{tabular}
\caption{Mean reward and standard deviation of executing the meta controller and the hierarchical planning algorithm over 5000 random instances of the increasing variance environment with two goal states.}
\label{table:sim_low_risk}
\end{table}

\section{Improving human decision-making by teaching automatically discovered planning strategies}\label{sec:exp-results}

\begin{figure}[ht]
    \centering
    \includegraphics[height=0.60\textwidth]{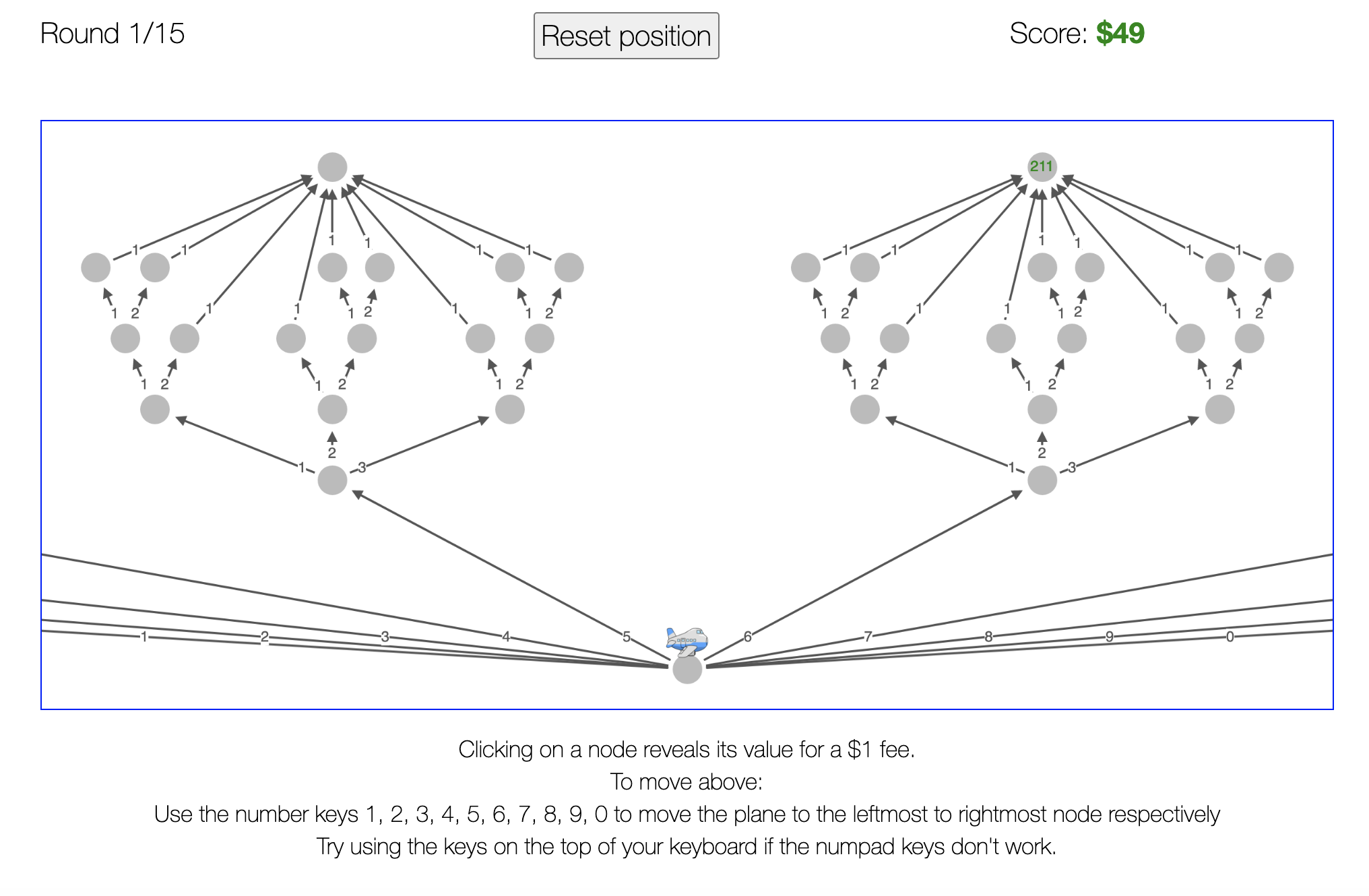}
    \caption{Screenshot of the Flight Planning task used to assess people's planning skills in Experiments 1. Participants can drag and zoom into the environment to show different portions.}
    \label{fig:FlightPlanningTask}
\end{figure}

Having shown that our method discovers planning strategies that achieve a super-human level of performance, we now evaluate whether we can improve human decision-making by teaching them the automatically discovered strategies. Building on the Mouselab-MDP paradigm introduced in Section~\ref{sec:MouselabMDP}, we investigate this question in the context of the Flight Planning task illustrated in Figure~\ref{fig:FlightPlanningTask}. Participants are tasked to plan the route of an airplane across a network of airports. Each flight gains a profit or a loss. Participants can find out how much profit or loss an individual flight would generate by clicking on its destination for a fee of \$1. The participant's goal is to maximize the sum of the flights' profits minus the cost of planning. Participants can make as few or as many clicks as they like before selecting a route using their keyboard. 

\begin{figure}[ht]
    \centering
    \includegraphics[height=0.60\textwidth]{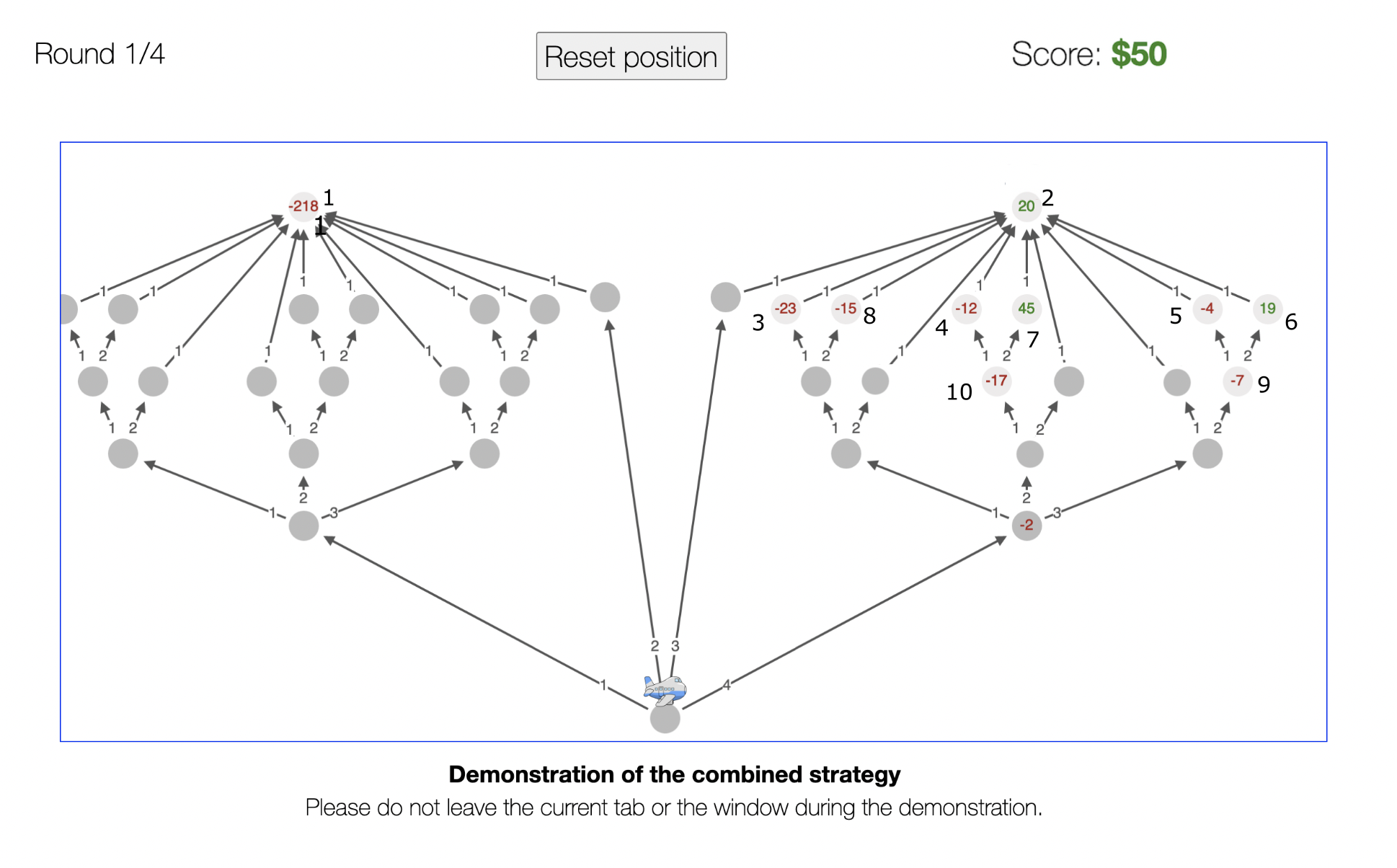}
    \caption{Screenshot of the cognitive tutor for demonstrating the non-hierarchical strategy  evaluated in Experiment 1. The numbers beside the node indicate the sequence in which the clicks were performed. }
    \label{fig:tutor}
\end{figure}

To teach people the automatically discovered strategies, we developed cognitive tutors (see Section~\ref{sec:cogtutbackground}) that shows people step-by-step demonstrations of what the optimal strategies for different environments would do to reach a decision (see Figure~\ref{fig:tutor}). In each step the strategy selects one click based on which information has already been revealed. At some point the tutor stops clicking and moves the airplane down the best route indicated by the revealed information. Moving forward we will refer to cognitive tutors teaching the hierarchical planning strategies discovered by hierarchical BMPS as \textit{hierarchical tutors} and refer to the tutors teaching the strategies discovered by non-hierarchical BMPS as \textit{non-hierarchical tutors}. 

To evaluate the effectiveness of these demonstration-based cognitive tutors, we conduct two experiments in which participants are taught the optimal strategies for flight planning problems equivalent to the two types of environments in which we evaluated our strategy discovery method in Section~\ref{sec:benchmarks}. To assess the potential benefits of the hierarchical tutors enabled by our new scalable strategy discovery method, these experiments compare the performance of people who were taught by hierarchical tutors against the performance of people who were taught by non-hierarchical tutors, the performance of people who were taught original feedback-based tutor for small environments \citep{CognitiveTutorsPNAS,CognitiveTutorsRLDM}, and the performance of people who practiced the task on their own. We developed the best version of each tutor possible given the limited scalability of the underlying strategy discovery method. The increased scalability of our new method enabled the hierarchical tutor to demonstrate the optimal strategy for the task participants faced whereas the other tutors could only show demonstrations on smaller versions of the task. We found that showing people a small number of demonstrations of the optimal planning strategy significantly improved their decision-making not only when the assumption underlying our method's hierarchical problem decomposition is met (Experiment~1) but also when it is violated (Experiment~2).

\subsection{Experiment 1: Teaching people the optimal strategy for an environment with increasing variance}\label{sec:hier_exp}

In the Experiment 1, participants were taught the optimal planning strategy for an environment in which $10$ final destinations can be reached through $9$ different paths comprising between 4 and 6 steps each (see Figure~\ref{fig:FlightPlanningTask}). The most important property of this environment is that the variance of available rewards doubles from each step to the next, starting from 5 in the first step. Therefore, in this environment, the optimal planning strategy is to first inspect the values of alternative goals, then commit to the best goal one could find, and then plan how to achieve it without ever reconsidering other goals.

%After participants in a  behavioral experiment with the Mouselab-MDP paradigm introduced in Section~\ref{sec:MouselabMDP}. 

% The size of the environment demonstrated was limited by the scalability of the algorithm's ability to find a solution within a reasonable amount of time (8 hours).

\begin{figure}[ht]
  \centering
  \includegraphics[height=0.60\textwidth]{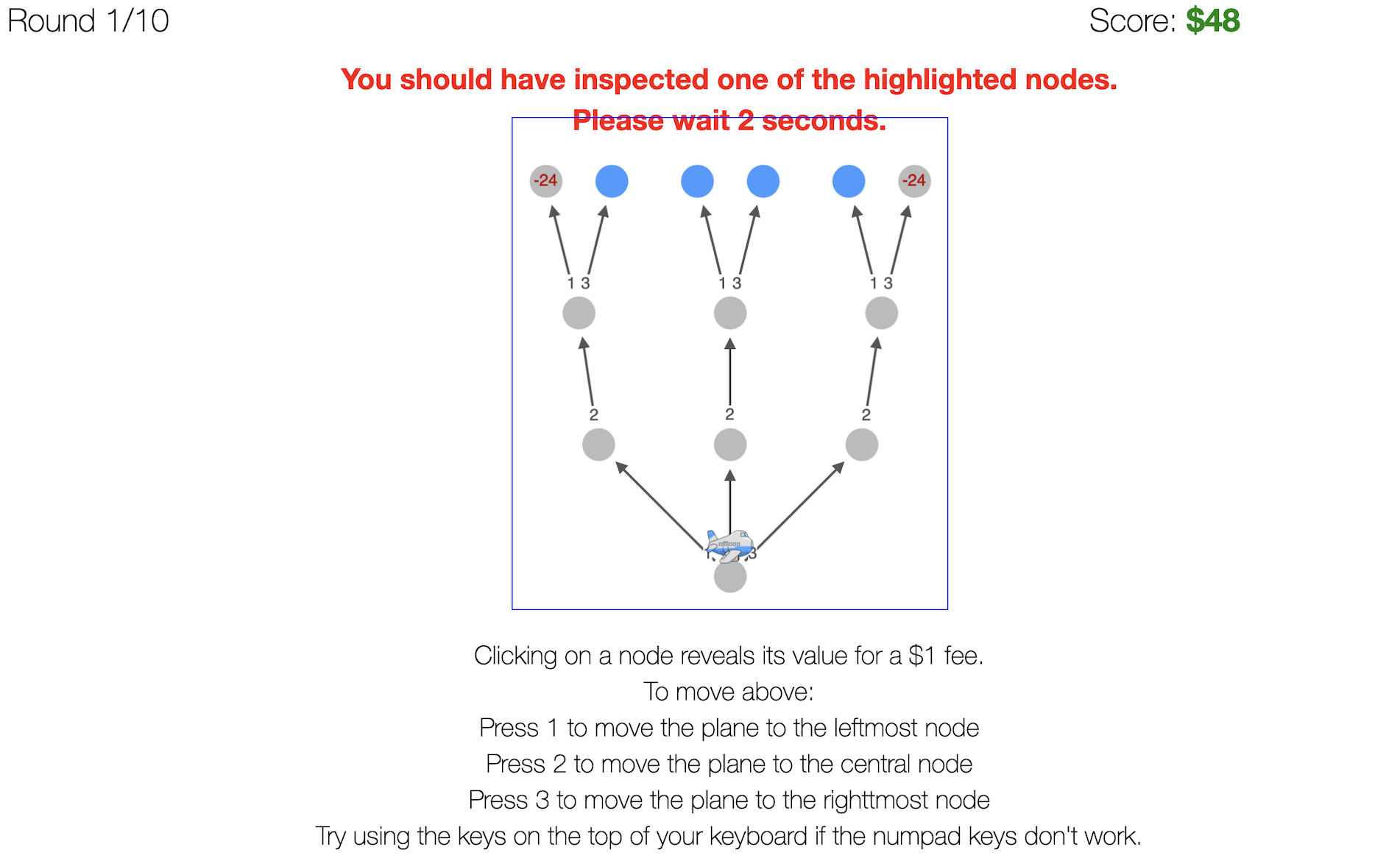}
  \caption{Screenshot of the feedback-based tutor against which we evaluated our scalable demonstration-based tutor. This tutor gives people feedback on the planning operations they perform in a smaller version of the environment. This small environment is the largest one for which optimal feedback can be computed with the currently available methods.}
  \label{fig:exp-312}
\end{figure}

\subsubsection{Methods}
We recruited $168$ participants on Amazon Mechanical Turk (average age $34.9$~years, range: $19$--$70$ years; $98$ female) \citep{litman2017turkprime}. Participants were paid \$$2.50$ plus a performance-dependent bonus (average bonus \$$2.86$). The average duration of the experiment was $46.9$~min. Participants could earn a performance-dependent bonus of 1 cent for every 10 points they won in the test trials.

All participants had to first agree to a consent form stating they were above 18, a US citizen residing in USA and fluent in English. After this, instructions about the range of possible rewards ($-250$ to $250$), cost of clicking (\$$1$), and the movement keys were presented. Then, participants went through $5$ trials to familiarize with the experiment. Following this, participants were either given additional $10$ practice trials or had $10$ trials with the cognitive tutor depending on their experimental condition. Finally, the participant was given $15$ test trials in the flight planning task with 10 possible final destinations illustrated in Figure~\ref{fig:FlightPlanningTask}. Participants started with $50$ points in the beginning of the test block.  

To evaluate the efficacy of cognitive tutors, participants were assigned to $4$ groups. In the experimental group participants were taught by the hierarchical tutor. The first control group was taught by the  non-hierarchical tutor. The second control group was taught by the feedback-based cognitive tutor \citep{CognitiveTutorsRLDM, CognitiveTutorsPNAS} illustrated in Figure~\ref{fig:exp-312}. The third control group practiced the Flight Planning task 10 times without feedback. The hierarchical tutor taught the strategy discovered by hierarchical BMPS algorithm discovered for the task participants had to perform in the test block. It first demonstrated $3$ trials with the goal selection strategy; it then showed three demonstrations of the goal-planning strategy; and finally presented $4$ demonstrations of the complete strategy combining both parts. The non-hierarchical tutor showed $10$ demonstrations of the strategy that non-hierarchical BMPS discovered for the largest version of the Flight Planning task it could handle (i.e., 2 goals instead of 10 goals). Computational bottlenecks confined the feedback-based tutor to a three-step planning task with six possible final destinations shown in Figure~\ref{fig:exp-312} \citep{CognitiveTutorsPNAS,CognitiveTutorsRLDM}. Participants received feedback on each of their clicks and their decision when the stop clicking as illustrated in Figure~\ref{fig:exp-312}. %The optimal strategy discovered by dynamic programming was to click on nodes with higher variance. Feedback was given to participants after performing an action to tell what action, 
When the participant chose a sub-optimal planning operation they were shown a message stating which planning operation the optimal strategy would have performed instead. In addition, they received a timeout penalty whose duration was proportional to how sub-optimal their planning operation had been.%the difference between the metalevel Q-value of the optimal planning operation and the metalevel Q-value of the chosen planning operation.

Counterbalanced assignment ensured that participants were equally distributed across four experimental conditions (i.e., $42$ participants per condition). %The expected score based on the clicks performed by the participants minus the cost of clicks was chosen as the metric for evaluation as shown in Figure~\ref{fig:ExpResults}.
%\paragraph{Exclusion Criteria}
To ensure high data quality, we applied a pre-determined exclusion criterion. We excluded $7$ participants who did not make a single click on more than half of the test trials because not clicking is highly indicative of speeding through the experiment without engaging with the task. 

\subsubsection{Results}

\begin{figure}[ht]
  \centering
  \includegraphics[height=0.35\textwidth]{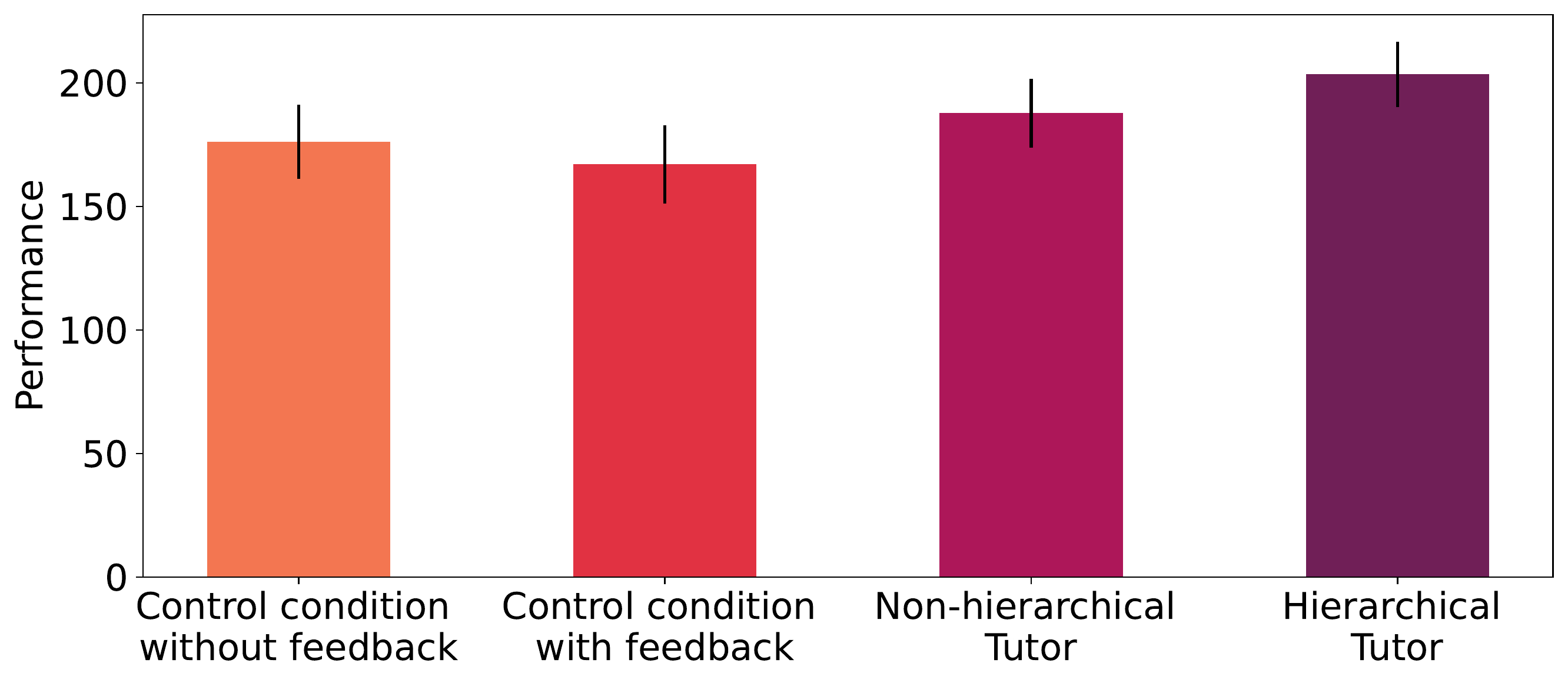}
  \caption{Performance  of  different  strategy  discovery  algorithms.}
  \label{fig:ExpResults}
\end{figure}

Figure~\ref{fig:ExpResults} shows the average performance of the four groups on the test trials. According to a Shapiro-Wilk test, participants' scores on the test trials were not normally distributed in any of the four groups (all $p<.001$). We therefore tested our hypothesis using non-parametric tests. To test if there are any significant differences between the groups in our repeated-measures design, we performed a Wald test.
We found that people's performance differed significantly across the four experimental conditions ($F=15.68$, $p=.001$). Planned pair-wise Wilcoxon rank-sum tests confirmed that teaching people strategies discovered by the hierarchical method significantly improved their performance (204.48 points/trial) compared to the control condition (177.36 points/trial, $p=.006$, $d=0.304$), the feedback-based cognitive tutor (167.02 points/trial, $p<.001$, $d=0.4$), and the non-hierarchical tutor ($p=.025$, $d=0.258$). By contrast, neither the feedback-based cognitive tutor ($p=.45$, $d=0.104$) nor the non-hierarchical cognitive tutor ($p=.67$, $d=0.045$) were more effective than letting people practice the task on their own. 
% teaching strategies discovered by the non-hierarchical method failed to improve people's performance (181.58 points/trial) compared to the control condition ($t(79)=0.18$, $p=0.6$, $d=0.045$) and the feedback-based cognitive tutor condition ($t(77)=1.35$, $p=0.2$, $d=0.148$)
% %, and also led to significantly lower performance than teaching strategies discovered by the hierarchical method ($t(79)=5.02$, $p=0.02$, $d=0.258$).
% Also the feedback based tutor's performance was not significantly better than the control condition ($t(78)=0.57$, $p=0.45$, $d=0.104$).
These results show that the hierarchical method is able to discover and teach the discovered strategy in environments in which previous methods failed.  

% \section{Limitations}
% The proposed hierarchical strategy discovery algorithm assumes that once a goal has been selected, there is no need to switch goals after midway during planning. Which means that it is able to discover effective for environments where the goal nodes have the most impact in final cumulative score, i.e, in which the goal nodes have the absolute highest value. To overcome this issue, we allowed for goal switching by introducing a meta-level controller as described in section~\ref{sec:meta}.

\subsection{Experiment 2: Teaching people the optimal strategy for a risky environment} \label{sec:meta_exp}

In Experiment~2, participants were taught the strategy our method discovered for the 8-step decision problem illustrated in Figure~\ref{fig:high_risk}. Critically, in this environment each path contains one risky node that harbors an extreme loss with a probability of 10\%. Therefore, the optimal strategy for this environment inspects the risky node while planning how to achieve the selected goal and then switches to another goals when it encounters a large negative reward on the path to the initially selected goal.

\subsubsection{Method}
%To evaluate the benefit of teaching the goal switching planning algorithm to people, we test our different strategies in an online experiment. 
To test whether our approach can also improve people's performance in environments with this more complex structure, we created two demonstration-based cognitive tutors that teach the strategies discovered by hierarchical BMPS with goal-switching and hierarchical BMPS without goal-switching, respectively, and a feedback-based tutor that teaches the optimal strategy for a 3-step version of the risky environment. 

%Especially, we compared people's performance when being taught the strategy discovered by our method when taking goal-switching into account to the strategy of a lessened version of our method where we disabled the goal-switching functionality.
This experiment used a Flight Planning Task that is analogous to the environment described in Section \ref{sec:metacontroller_eval} (see Figure~\ref{fig:high_risk}). Specifically, the environment comprises 4 goal nodes and 60 intermediate nodes (i.e., 15 per goal). Although each goal can be reached through multiple paths all of those paths lead through an unavoidable node that  has a 10\% risk of harboring a large loss of -1500. The aim of this experiment is to verify that we are still able to improve human planning even when environment requires a more complex strategy that occasionally switches goals during planning. To test this hypothesis we showed the participants demonstrations of the strategy discovered by our method in the experimental condition, and compared their performance to the performance of three control groups. The first control group was shown demonstrations of the strategy discovered by the version of our method without goal switching; the second control group discovered their own strategy in five training trials; the third control group practiced planning on a three-step task with a feedback-based tutor (see Section \ref{sec:cogtutbackground}) \citep{CognitiveTutorsPNAS, CognitiveTutorsRLDM}. The environment used by the feedback-based tutor mimicked the high-risk environment. To achieve this we changed the reward distribution of the intermediate nodes so that there is a 10\% chance of a negative reward of -96, a 30\% chance of -4, a 30\% chance of +4, and a 30\% chance of +8. We then recomputed the optimal feedback using dynamic programming \citep{CognitiveTutorsPNAS,CognitiveTutorsRLDM}.

We recruited 201 participants (average age $34.01$ years, range:19–70 years; $101$ female) on Amazon's Mechanical Turk \cite{litman2017turkprime} over three consecutive days. All but two of them completed the assignment. Applying the same pre-determined exclusion criterion as we used in Experiment 1 (i.e., excluding participants who do not engage with the environment in more than half of the test trials) led to the exclusion of 30 participants (15\%), leaving us with $169$ participants. Participants were paid 1.30\$ and a performance dependent bonus of up to 1\$. The average bonus was 0.56\$ and the average time of the experiment was 16.28 minutes. Participants were randomly assigned to one of four experimental conditions determining their training in the planning task. All groups were tested in five identical test trials. The data was analysed using the robust \texttt{f1.ld.f1} function of the \texttt{nparLD} R package \citep{nparld}. On each trial the participant's score was calculated as the expected reward of the path they chose minus the the cost of the clicks they had made\footnote{As in the simulations, the cost per click was 10.}. 

\subsubsection{Results}
The results of the experiment are summarized in Table \ref{table:exp_high_risk}. Since the Shapiro-Wilk test shows that none of the four conditions are normally distributed ($p<.001$ for all), we again use the non-parametric Wald test to evaluate the data. The Wald test shows significant differences between the four groups ($F=62.6$, $p<.001$). Pairwise robust post-hoc comparisons show that participants trained with demonstrations of the strategy discovered by hierarchical BMPS with goal-switching significantly outperform all other conditions: participants trained by purely hierarchical demonstrations without goal switching ($p<.001$, $d=0.72$), participants who did not receive demonstrations ($p<.001$, $d=0.51$), and participants who had practiced planning with optimal feedback on a smaller analogous environment ($p<.001$, $d=0.53$). The performance of the three control groups was statistically indistinguishable (all $p \geq 0.18$). Participants trained by purely hierarchical demonstrations did not perform significantly better than participants that trained with optimal feedback ($p=.18$, $d=0.11$) or participants that did not receive demonstrations ($p=.56$, $d=0.07$). Additionally there was no significant difference between the optimal feedback condition and the no demonstration condition ($p=.6$, $d=0.03$). 

%\subsubsection{Discussion}
The results of this experiment show that we can significantly improve human decision-making by showing them demonstrations of the automatically discovered hierarchical planning strategy with goal-switching. This is a unique advantage of our new method because none of the other approaches was able to improve people's decision-making in this large and risky environment.
By comparing human performance to the optimal performance of our algorithm in the same environment (see Table \ref{table:sim_high_risk}) we can see that even though we were able to improve human performance, participants still did not fully understand the strategy based on the demonstrations alone. This reveals the limitations of teaching planning strategies purely with demonstrations, especially for more complex strategies. Improving upon the pedagogy of our purely demonstration-based hierarchical tutor is an important direction for future work.

\begin{table}[ht]
\centering
\begin{tabular}{l|ccc}
\toprule
Algorithm       & N    & Reward & Std    \\ 
\midrule
No demonstration & 42 & -103.31 & 173.69 \\
Goal-switching demonstration    & 45 &    -26.71   & 121.87  \\ 
Hierarchical demonstration & 39 &   -94.41    & 51.75 \\
Feedback tutor & 43 & -108.49 & 179.64 \\
\bottomrule
\end{tabular}
\caption{Experimental results of teaching people automatically discovered planning strategies for the high risk environment shown in Figure~\ref{fig:high_risk}. For each condition we report the number of participants, the mean expected reward and the standard deviation of the mean expected reward.}
\label{table:exp_high_risk}
\end{table}

\section{General Discussion}

%\js{I think this section is too long. We could either shorten it or split it into subsections.}

%The design of efficient algorithms could, in principle, be partly automated by using machine learning methods for automatic strategy discovery. But previous strategy discovery methods lacked scalability.
%In our quest to design scalable strategy discovery methods, it is seen that existing strategy discovery methods lack scalability which is exemplified with the exponential increase in run time of BMPS with the domain size.
% of the domain.

%context: problem and previous work
To make good decisions in complex situations people and machines have to use efficient planning strategies because planning is costly. Efficient planning strategies can be discovered automatically. But computational challenges confined previous previous strategy discovery methods to tiny problems. To overcome this problem, we devised a more scalable machine learning approach to automatic strategy discovery.
%content: summary of our approach and our findings
To overcome this problem, we devised a more scalable machine learning approach to automatic strategy  discovery. The central idea of our method is to decompose the strategy discovery problem into discovering goal-setting strategies and discovering goal achievement strategies. In addition, we made a substantial algorithmic improvement to the state-of-the-art method for automatic strategy discovery \citep{callaway2017learning} by introducing the tree-contraction method.
We found that this hierarchical decomposition of the planning problem, together with our tree contraction method, drastically reduces the time complexity of automatic strategy discovery 
% (Figure~\ref{fig:eval-merged}b) 
without compromising on the quality of the discovered strategies 
% (Figure~\ref{fig:eval-merged}a) 
in many cases. Furthermore, by introducing the tree contraction method 
% allows us to 
we have extended the set of environment structures that automatic strategy discovery can be applied to from trees to directed acyclic graphs.
%makes it possible to discover planning strategies for environments that include undirected cycles
%enables solving new environment types by being able to resolve undirected cycles in the tree structure. Crucially, it extends the range of solvable problem structures
% The range of possible problems we can now solve is therefore extended 
%from trees to directed acyclic graphs (DAG).
 %conclusion: this is a significant advance
These advances significantly extend the range of strategy discovery problems that can be solved by making the algorithm faster, more scalable, and applicable to environments with more complex structure. 
This is an important step towards discovering efficient planning strategies for real-world problems.

%context: people struggle to make good decisions in large environments, people's decision strategies can be improved, but so far this is only possible for toy problems
Recent findings suggests that teaching people automatically discovered efficient planning strategies is promising way to improve their decisions \citep{CognitiveTutorsPNAS,CognitiveTutorsRLDM}. Due to computational limitations this approach was previously confined to sequential decision problems with at most three steps \citep{CognitiveTutorsPNAS,CognitiveTutorsRLDM}. 
%content: cognitive tutors and empirical findings
The strategy discovery methods developed in this article make it possible to scale up this approach to larger and more realistic planning tasks. As a proof-of-concept, we showed that our method makes it possible to improve people's decisions in planning tasks with up to 7 steps and up to 10 final goals. %Additionally, 
% we verified that we are able to 
%our method can effectively teach the discovered decision strategies
% our method discovered 
We evaluate the effectiveness of showing people demonstrations of the strategies discovered by our method in two separate experiments where the environments were so large that previous methods were unable to discover planning strategies within a time budget of eight hours. Thus, the best one could to at training people with previous methods was to construct cognitive tutors that taught people the optimal strategy for a smaller environment with similar strategy or having people practice without feedback. Evaluating our method against these alternative approaches we found that our approach was the only one that was significantly more beneficial than having people practice the task on their own.
%compared the effectiveness of showing people demonstrations of the strategies discovered by our method against the effectiveness of the best cognitive tutors that could be created with previous strategy discovery methods or having people practice the task on their own. We compared these approaches
%The strategy discovered by our method was taught to participants using demonstrations and improved their performance compared to control groups using self-training or feedback-based training on smaller environments.
%conclusion
To the best of our knowledge, this makes our algorithm the only strategy discovery method that can improve human performance on sequential decision problems of this size. This suggests that our approach makes it possible to leverage reinforcement learning to improve human decision-making in problems that were out of reach for previous intelligent tutors.
Our method's hierarchical decomposition of the planning problem exploits that people can typically identify potential mid- or long-term goals that might be much more valuable than any of the rewards they could attain in the short run. This corresponds to the assumption that the rewards available in more distant states are more variable than the rewards available in more proximal states.
%content: 
When this assumption is satisfied, our method discovers planning strategies much more rapidly than previous methods and the discovered strategies are as good as or better than those discovered with the best previous methods.
%rapidly discovers by our method perform as well as the strategies discovered by the much slower original BMPS method that was previously the state-of-the-art.
% (see Section \ref{sec:hierarchical_eval}). 
When this assumption 
% assumption of increasing variance 
is violated, the goal switching mechanism of our method can compensate for that mismatch. This allows the discovered strategies to perform almost as well as the strategies discovered by BMPS. Our method relies on this mechanism more the more strongly its assumption is violated. In doing so, it automatically trades off its computational speed-up against the quality of the resulting strategy.
%comes at the cost of diminishing the computational speed-up relative to the original version of BMPS. %Goal switching enables our method to avoid bad decisions in complex risky environments at the expense of additional computational costs.
%avoids large losses by switching goals when the discovered nodes show a lower reward than expected. This leads to a large performance gain compared to the lessened, purely hierarchical version of our method
% (see Section \ref{sec:metacontroller_eval}) 
%on certain environments, and helps mitigate the trade-off between performance and efficiency caused by incomplete environment evaluation.
% not evaluating the complete environment.
% conclusion: robustness to violations of the assumed structure
This shows that our method is robust to violations of its assumptions about the structure of the environment; it exploits simplifying structure only when it exists.

%relationship to previous work
%Our work goes beyond recent work on goal-conditioned planning by \citet*{nasiriany2019planning} 
% Nasiriany et al. (2019) 
%and \citet*{pertsch2020long}. 
% Pertsch et al. (2020). 
%We apologize for our previous ignorance of this literature. We will cite this impressive work and clarify how our work differs from it. 
%Fortunately for us, 
Some aspects of our work share similarities with recent work on goal-conditioned planning \citep{nasiriany2019planning,pertsch2020long}, although the problem we solved is conceptually different.
% \cite{nasiriany2019planning} % Nasiriany et al. (2019)
% and \cite{pertsch2020long}. 
% One difference is that 
For comparison, both aforementioned methods 
% \cite{nasiriany2019planning, pertsch2020long}
optimize the route to a given final location, whereas our method learns a strategy for solving sequential decision problems where the strategy chooses the final state itself.
% has to be chosen by the algorithm itself. 
Furthermore, while \citet{nasiriany2019planning} specified a fixed strategy for selecting the sequence of goals, our method learns such a strategy itself. Critically, while policies learned by \citet{nasiriany2019planning} select physical actions (e.g., move left), the metalevel policies learned by our method select planning operations (i.e., simulate the outcome of taking action $a$ in state $s$ and update the plan accordingly). 
Finally, our method explicitly considers the cost of planning to find algorithms that achieve the optimal trade-off between the cost of planning and the quality of the resulting decisions.

%\section{Limitations and scope for future work}
%\begin{itemize}
%    \item Goal discovery not part of algorithm
 %   \item Performance is dependent on knowledge of prior distribution. Past work in BMPS \cite{KemturJain2020} indicate robustnes but robust analysis for hierarchical BMPS to be done
 %   \item If cumulative variance of intermediate path is greater than goal nodes, performance degrades
%    \scon{Should we move Table 4 and the analysis to this section?}
%\end{itemize}

%limitations
%context:
Our method's scalability has its price. 
%content:
Since our approach decomposes the full sequential decision problem into two sub-problems (goal-selection and goal-planning), its accuracy can be limited by the fact that it never considers the whole problem space at once.
%Our method is inherently limited by the fact that planning only in a subset of the complete problem can never be as accurate as planning in the whole problem space.
This is unproblematic when the environment's structure matches our method's assumption that the rewards of potential goals are more variable than more proximal rewards. But it could be problematic when this assumption is violated too strongly. We 
% were also able to 
mitigated this potential problem by allowing the strategy discovery algorithm to switch goals. Even with this adaptation, the discovered strategy is not optimal in all cases: Since the representation of the alternative goal reward is defined as its average expected reward, the algorithm will only switch goals if the current goal's reward is below average. However, if the current goal's expected return is above average, the discovered strategy will not explore other goals even when that would lead to a higher reward. %We tried circumventing this issue by making the alternative goal reflective of the distribution over possible rewards in the alternative sub-tree, but did not find an increase in performance with this approach. 
%conclusion
On balance, we think that the scalability of our method to large environments outweighs this minor loss in performance.
% the small loss in performance compared to previous methods is outweighed by being able to solve much larger problems that previous methods are unable to solve.

%directions for future work
%context
The advances presented in this article open up many exciting avenues for future work. 
%content
For instance, our approach could be extended to plans with potentially many levels of hierarchically nested subgoals. Future work might also extend our method so that any state can be selected as a goal.
%to our method lies in the definition of goal states. 
In its current form, our algorithm always selects only the environment's most distant states (leaf nodes) as candidate goals. 
% A possible extension of this is to 
Future versions might allow the set of candidate goals to be chosen more flexibly such that some leaf nodes can be ignored and some especially important intermediate nodes in the tree can be considered as potential sub-goals. A more flexible definition and potentially a dynamic selection of 
% which nodes qualify as 
goal nodes could increase the strategy discovery algorithm's performance, and possibly allow us to solve a wider range of more complex problems. This would mitigate limitations of the increasing variance assumption by considering all potentially valuable states as (sub)goals regardless of where they are located.

%The proposed algorithms is applicable to more complex environments. Optimal planning strategies have been obtained for environments having a cyclic path. In the cyclic environment, the node with the greatest depth in the cycle is considered as a goal state.
%Furthermore, \red{ADD LINE ABOUT STOCHASTIC TRANSITION}
%\scon{Running simulation for 2-36 environment with 10\%, 20\%, 30\%, 50\% stochasticity} \lovis{We can not solve environments with cyclic paths as far as I know. Is the stochastic transition still relevant?}

%Future work will evaluate the performance of automatically discovered hierarchical planning strategies against some of the best hierarchical planning algorithms designed by AI researchers \cite{Kaelbling2010,Wolfe2010}.
%Furthermore, we have studied the range of problems where hierarchical decomposition would be useful.
%We found that its usefulness is linked to the usefulness of performing goal node computations. 

%context
The advances reported in this article have potential applications in artificial intelligence, cognitive science, and human-computer interaction. 
%content
First, since the hierarchical structure exploited by our method exists in many real-world problems, it may be worthwhile to apply our approach to discovering planning algorithms for other real-world applications of artificial intelligence where information is costly. This could be a promising step toward AI systems with a (super)human level of computational efficiency.
% We found that our hierarchical decomposition is useful for a wide range of planning problems -- especially when the values of candidate goals are more variable than the 
% % costs of the paths that lead there.
% path costs for achieving these goals.
% In real world problems, the benefit for a possible action is more uncertain as the planning step increases. 
%In real world long-horizon problems, the pay-off between goals is more variable than the pay-off of intermediate steps. 
%The hierarchical structure exploited by our method is present in many real-world problems.%, including the challenge of choosing a career. 
% Going back to the student's dilemma of becoming a computer scientist or a janitor, the impact on someones live from the choice of being a computer scientist or a janitor is much larger than the impact of the selection of subjects in school. Hence, the variance of the goal nodes would, most likely, be higher than the variance of the possible paths to reach the goal.
%Thus, the hierarchical decomposition introduced in this article might prove useful for the automatic algorithm discovery for lifelike problems. 
Second, our method also enables cognitive scientists to scale up the resource-rational analysis methodology for understanding the cognitive mechanisms of decision-making \citep{LiederGriffiths2020} to increasingly more naturalistic models of the decision problems people face in real life. 
Third, future work will apply the methods developed in this article to train and support people in making real-world decisions they frequently face. Our approach is especially relevant when acquiring information that might improve a decision is costly or time consuming. This is the case in many real-world decisions. For instance, when a medical doctor plans how to treat a patient's symptoms acquiring an additional piece of information might mean ordering an MRI scan that costs \$1000. Similarly, a holiday planning app would have to be mindful of the user's time when deciding which series of places and activities the user should evaluate to efficiently plan their road trip or vacation.
%when asking them to evaluate how much they would enjoy different places they might visit on a road trip or various activities they might do at various times throughout their vacation. 
Similar tradeoffs exist in project planning, financial planning, and time management. Furthermore, our approach can also be applied to support the information collection process of hiring decisions, purchasing decisions, and investment decisions. Our approach could be used to train people how to make such decisions with intelligent tutors \citep{CognitiveTutorsPNAS,CognitiveTutorsRLDM}. Alternatively, the strategies could be conveyed by decision support systems that guide people through real-life decisions by asking a series of questions. In this case, each question the system asks would correspond to an adaptively chosen information gathering operation. 
%conclusion: closing sentence/paragraph: optimistic outlook or positive conclusion
In summary, the reinforcement learning method developed in this article is an important step towards intelligent systems with a (super)human-level computational efficiciency, understanding how people make decisions, and leveraging artificial intelligence to improve human decision-making in the real world. At a high level, our findings support the conclusion that incorporating cognitively-informed hierarchical structure into reinforcement learning methods can make them more useful for real-world applications.

\section*{Declarations}

\paragraph{Funding}
This project was funded by grant number CyVy-RF-2019-02 from the Cyber Valley Research Fund.

\paragraph{Conflicts of interest/Competing interests}
The authors declare that the have no conflicts of interest or competing interests.

\paragraph{Availability of data and material (data transparency)}
All materials of the behavioral experiments we conducted are available at \url{https://github.com/RationalityEnhancement/SSD_Hierarchical/master/Human-Experiments}.
Anonymized data from the experiments is available at \url{https://github.com/RationalityEnhancement/SSD_Hierarchical/master/Human-Experiments}.

\paragraph{Code availability (software application or custom code)}
The code of the machine learning methods introduced in this article is available at \url{https://github.com/RationalityEnhancement/SSD_Hierarchical}.

\paragraph{Ethics approval}
The experiments reported in this article were approved by the IEC of the University of T\"ubingen
under IRB protocol number 667/2018BO2 (``Online-Experimente \"uber das Erlernen von Entscheidungsstrategien'').

\paragraph{Consent to participate (include appropriate statements)}
Informed consent was obtained from all individual participants included in the study.

\paragraph{Consent for publication}
Not applicable.

%\end{ack}

% \bibliographystyle{spmpsci}      % mathematics and physical sciences
% \bibliographystyle{spphys}       % APS-like style for physics
\bibliography{references}  % name your BibTeX data base
\newpage
\input{appendix}

\end{document}

%% file: appendix.tex
\appendix
\renewcommand{\thesection}{A.\arabic{section}}
\renewcommand{\thefigure}{A\arabic{figure}}
\section*{\Large{Appendix}}  % This is NOT part of the template!

\section{VOC features}
\begin{figure}[h]
  \centering
  \includegraphics[height=0.25\textwidth]{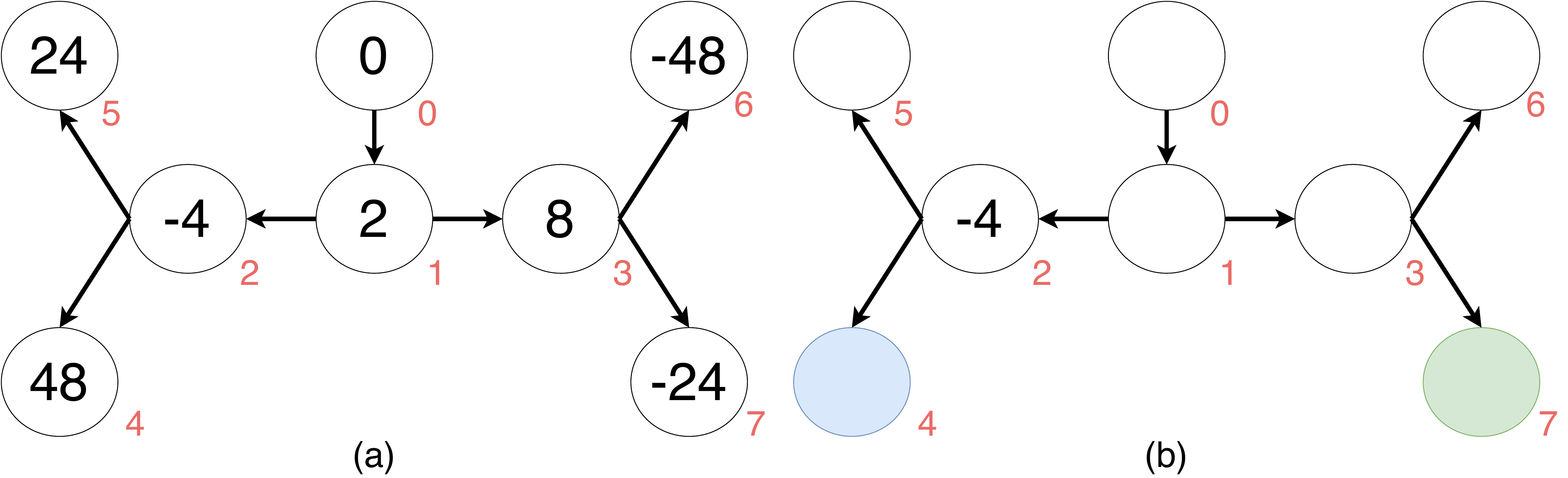}  
  \caption{(a) Simplified example environment with the rewards beneath each node marked in the figure. In the beginning, the rewards are hidden. Metalevel actions reveal the corresponding reward of a node. (b) State of the environment. Two metalevel actions are being considered which reveal the corresponding nodes as denoted by the blue (4) and green (7) color.}
  \label{fig:voc-env}
\end{figure}

%\lovis{I'm not sure the examples on the simplified MDP are necessary}
%\scon{Need to define Mouselab MDP, so this should be after subsection "The Mouselab-MDP paradigm".}
The features used to approximate the value of information can be explained using a simplified Mouselab-MDP environment. The ground truth values of each node is depicted in Figure~\ref{fig:voc-env}(a). The rewards are sampled from a Categorical equiprobable distribution of $\{-2,-1,1,2\}$, $\{-8,-4,4,8\}$ and $\{-48,-24,24,48\}$ for nodes of depth $1$, $2$ and $3$ respectively. In this example, the value of information for two metalevel action are considered, marked in Figure~\ref{fig:voc-env}(b). 

To compute the VOC, the possible values of a subset of nodes are considered and the cumulative reward accumulated from the maximal path is computed to find the expected return if the node values were known. Subtracting this from the cumulative reward from the maximal path given the current belief state gives the value of information of knowing performing the metalevel action. The greater the difference in the two quantities imply greater information is possibly gained for performing the computation.

In case of myopic VOC computation, the subset of nodes considered is just the node whose reward would be revealed by the metalevel action. $\mathrm{VPI}$ considers the entire subset of nodes in the environment. $\mathrm{VPI_{sub}}$ considers all nodes lying on paths which pass through the node revealed by the computation. For example, when considering the $\mathrm{VPI_{sub}}$ for the computation which reveals the node marked in blue in Figure~\ref{fig:voc-env}(b), nodes $1$, $2$ and $4$ would be considered. Whereas, nodes $1$, $3$ and $7$ would be considered for the computation corresponding to revealing the value of the green node. 

\section{Time complexity upper bound analysis}\label{sec:time}
In this section we analyse the computational time upper bound of the methods that we used. For simplicity, in the hierarchical case, we assume that goal switching is not possible. That means that the high level controller would run once, followed by the low level controller.
To ensure readability, we explicitly define the notations used in this section and throughout the paper:
\begin{itemize}
    \item $N$: number of intermediate nodes to goal
    \item $M$: number of goal states
    \item $B$: number of bins to discretize a continuous probability distribution
    \item $\mathrm{RUN}$: number of unrevealed nodes relevant to compute feature
\end{itemize}

The hierarchical decomposition reduced the time complexity of the myopic strategy discovery problem from $O((M \cdot N)^2 \cdot B)$ to $O((M^2 + N^2) \cdot B)$. For the BMPS algorithm, the hierarchical structure reduces the computational time upper bound from $O((M \cdot N)^2 \cdot B^{N})$ to $O(M^2 \cdot B + B^{M}  + N^2 \cdot B^{N})$. The reduction in the upper bound implies that algorithms that use hierarchical structure are scalable to more complex environments.

As discussed in
Section~\ref{sec:metaMDP}, metalevel actions are selected based on maximising the approximate VOC.
At each step, a metalevel action that maximizes the approximate VOC is chosen. Selection of a metalevel action converts the probability distribution of the chosen node to a Dirac delta function concentrated at the revealed node value. The calculation of VOI features for continuous probability distributions requires computations of multiple cumulative distribution functions (CDF). In general, this procedure is computationally expensive and an approximation of it inherently requires discretization. Hence, the probability density function (PDF) of a continuous probability distribution associated with a node in the environment is discretized into $B$ bins. As the number of bins $B$ increases, the discrepancy between the approximation and the true PDF/CDF decreases and it shrinks to $0$ as $B \to \infty$ at the cost of higher computation cost.

The number of relevant unrevealed nodes ($\mathrm{RUN}$) is the count of unrevealed nodes relevant to compute a VOI feature. 
% Here, we state the number of relevant nodes and how they affect the computation time for \red{different types of algorithms}.
% This number
The $\mathrm{RUN}$
varies for different algorithm features and directly affects the time required to compute
% the values of the features.
their values for a given state-action pair. 
For calculating the approximate VOC, the number of possible values to be compared
% would be equal to 
is $B^{\mathrm{RUN}}$. Each possible value set requires $\alpha \in \mathbb{R}_{\neq 0}$ time to compute the highest cumulative return from all the possible outcomes of the set. Hence, the  time
% taken
required to calculate the approximate VOC scales with $\alpha \cdot B^{\mathrm{RUN}}$. %\red{$\mathrm{RUN}$-tuple bin values}.
For myopic VOC estimation, the number of relevant unrevealed nodes is 1. 
For BMPS, the number of relevant nodes for each of the VOI features $\mathcal{F}=\lbrace \mathrm{VOI_{1}}, \mathrm{VPI}, \mathrm{VPI_{sub}} \rbrace$ is different. The VOI$_{1}$ feature requires the least time for computation since its value depends on only one node in the environment.
On the other hand, the most time-consuming calculation is for the $\mathrm{VPI}$ feature since its value depends on the all nodes in the environment. The $\mathrm{VPI_{sub}}$ feature considers values of all paths that pass through a node evaluated by the metalevel action. Hence, the most time consuming calculation of $\mathrm{VPI_{sub}}$ is
% would be
for metalevel actions that correspond to the goal node.
% all nodes are relevant .
Speaking about algorithmic complexity in terms of the big-O notation, it takes $O(B)$ time to calculate the myopic VOC value for a given
% action in a given state.
state-action pair.
On the contrary, it takes $O(B^{\mathrm{RUN}})$ time to calculate the $\mathrm{VPI_{sub}}$ value for a given
% action in a given state.
state-action pair.

\begin{table}[]
    \centering
    \begin{tabular}{lccc}
        \toprule
        Strategy Discovery Algorithm & Feature & $O(\mathrm{RUN})$ & $O$ \\
        \midrule
        \multirow{5}{*}{Hierarchical BMPS} & $\mathrm{VOI_{1}^{H}}$ & $1$ & $B$ \\
        % & $\mathrm{VPI^{H}}$ & $M$ & $O$($B^{M}$) \\
        % & $\mathrm{VOI_{1}^{L}}$ & $1$ & $O$($B$) \\
        % & $\mathrm{VPI_{sub}^{L}}$ & $N$ & $O$($B^{N}$) \\
        % & $\mathrm{VPI^{L}}$ & $N$ & $O$($B^{N}$) \\
        % \midrule
        % \multirow{2}{*}{Hierarchical greedy myopic VOC}
        % & $\mathrm{VOI_{1}^{H}}$ & $1$ & $O$($B$) \\
        % & $\mathrm{VOI_{1}^{L}}$ & $1$ & $O$($B$) \\
        % \midrule
        % \multirow{3}{*}{Non-hierarchical BMPS}
        % & $\mathrm{VOI_{1}}$ & $1$ & $O$($B$) \\
        % & $\mathrm{VPI_{sub}}$ & $M \cdot (N+1)$ & $O$($B^{M \cdot (N+1)}$) \\
        % & $\mathrm{VPI}$ & $M \cdot (N+1)$ & $O$($B^{M \cdot (N+1)}$) \\
        % \midrule
        % Non-hierarchical greedy myopic VOC & $\mathrm{VOI_{1}}$ & $1$ & $O$($B$) \\
        
        & $\mathrm{VPI^{H}}$ & $M$ & $B^{M}$ \\
        & $\mathrm{VOI_{1}^{L}}$ & $1$ & $B$ \\
        & $\mathrm{VPI_{sub}^{L}}$ & $N$ & $B^{N}$ \\
        & $\mathrm{VPI^{L}}$ & $N$ & $B^{N}$ \\
        \midrule
        \multirow{2}{*}{Hierarchical greedy myopic VOC}
        & $\mathrm{VOI_{1}^{H}}$ & $1$ & $B$ \\
        & $\mathrm{VOI_{1}^{L}}$ & $1$ & $B$ \\
        \midrule
        \multirow{3}{*}{Non-hierarchical BMPS}
        & $\mathrm{VOI_{1}}$ & $1$ & $B$ \\
        % & $\mathrm{VPI_{sub}}$ & $M \cdot (N+1)$ & $B^{M \cdot (N+1)}$ \\
        & $\mathrm{VPI_{sub}}$ & $N$ & $B^{N}$ \\
        % & $\mathrm{VPI}$ & $M \cdot (N+1)$ & $B^{M \cdot (N+1)}$ \\
        & $\mathrm{VPI}$ & $M \cdot N$ & $B^{M \cdot N}$ \\
        \midrule
        Non-hierarchical greedy myopic VOC & $\mathrm{VOI_{1}}$ & $1$ & $B$ \\
        \bottomrule
    \end{tabular}
    \caption{Asymptotic time to compute a feature for hierarchical and non-hierarchical strategy discovery algorithms}
    \label{tab:runtimeComparison}
\end{table}

The maximum amount of
% upper bound on the
computational time to calculate the approximate VOC directly depends on
% would be resulted by
the selection of all possible metalevel actions, for which we prove upper bounds for both the non-hierarchical (in Section~\ref{sec:timeNonH}) and hierarchical strategy discovery problem (in Section~\ref{sec:timeH}).
% In this analysis, we find the upper bound for the maximum time.

\section{Time complexity of the non-hierarchical strategy discovery problem}  \label{sec:timeNonH}

% In the first step of the
In the setting of the non-hierarchical strategy discovery problem, the metalevel policy has to initially select the best metalevel action from $M \cdot (N+1) + 1$ possible actions. This selection requires $M \cdot (N+1)$ VOI feature computations. Since computation of $\mathrm{VPI}$ is required only once for a given state, the next computationally most-expensive
% most time-consuming
feature is $\mathrm{VPI_{sub}}$, which is computed for each possible action. From all possible actions, the ones that demand most of the computational time to compute $\mathrm{VPI_{sub}}$ are the actions that correspond to the goal nodes. In this case, the computation of $\mathrm{VPI_{sub}}$ takes
% The action which takes the most time to compute  is for actions corresponding to the goal nodes,
$O(B^{N})$ time.
% The number of possible actions decreases linearly with each metalevel action selection.

% Considering that
In general, if there are $M$
% possible
goals in the environment and each goal consists of $N + 1$ nodes (i.e. $N$ intermediate nodes + $1$ goal node), the maximum number of metalevel actions
% that can be
performed including the termination action is $M \cdot (N + 1) + 1$.

In the non-hierarchical BMPS strategy discovery problem, the computational time upper bound to perform all metalevel actions and terminate is
\begin{align}
    % \textrm{Max Time} = 
      \sum_{i=0}^{M \cdot (N+1) + 1}\left[M \cdot (N+1) - i \right] \cdot O (B^{(N+1)})
    & = \sum_{i=0}^{M \cdot (N+1)}\left[M \cdot (N+1) - i \right] \cdot O (B^{N})  \\
    & = O((M \cdot N)^{2} \cdot B^{N})
\end{align}

For the greedy myopic strategy discovery algorithm, the number of relevant nodes ($\mathrm{RUN} = 1$) reduces the second term in the equation to $O(B)$.
% Therefore, the total maximum time can be is simplified to
In this case, the computational time upper bound to perform all metalevel actions and terminate is
\begin{align}
%   \textrm{Max Time} =
      \sum_{i=0}^{M \cdot (N+1) + 1}[M \cdot (N+1) + 1 - i] \cdot O (B)
    & = \sum_{i=0}^{M \cdot (N+1)}[M \cdot (N+1) + 1 - i] \cdot O (B)  \\
    & = \frac{(M \cdot (N+1) + 1) \cdot (M \cdot (N+1) + 2)}{2} \cdot O (B)  \\
    & = O((M\cdot N)^2 \cdot B)
\end{align}

\section{Time complexity of the hierarchical strategy discovery problem} \label{sec:timeH}

In the setting of the hierarchical strategy discovery problem, the action space shrinks severely for each metalevel action selection. During the goal-setting phase of the procedure, the number of high-level actions including the high-level policy termination is $M + 1$. The selection of a high-level action, requires the $1$ computation of $\mathrm{VPI^H}$ and at most $M$ computations of $\mathrm{VOI_{1}^{H}}$.
Therefore, the computational time upper bound for this phase is
\begin{equation}
    % \textrm{Time for high level policy} = 
        \sum_{i=0}^{M} \left[ (M- i) \cdot O (B) + O (B^{(M - i)}) \right]
      = O(M^{2} \cdot B) + \sum_{i=0}^{M} O (B^{(M - i)})
      = O(M^{2} \cdot B + B^{M})
\end{equation}

Similarly, during the goal-achievement phase of the algorithmic procedure, the number of low-level metalevel actions for each goal is $N + 1$. The most time consuming feature calculated in the goal-achievement phase is $\mathrm{VPI_{sub}^{L}}$.
% a maximum of $N+1$ nodes are explorable for each goal,
Therefore, the computational-time upper bound for this phase per goal is
% leading to the maximum time taken for the low level policy to be:
\begin{equation}
    % \textrm{Time for low level policy} = 
      \sum_{i=0}^{N+1}(N+1- i) \cdot O (B^{(N + 1 - i)})
    = O (N^2 \cdot B^{N)})
    %\sum_{i=0}^{N}(N+1- i) \cdot O (B^{(N + 1 - i)})
\end{equation}

To calculate the
% total
upper bound for the hierarchical strategy discovery algorithm, the
% maximum
combined computational time for the high-level and low-level policy is the sum of the computational time consumed on both levels independently. Additionally, since the value of a goal node is not required 
% to find the best strategy to achieve a goal,
in the goal-achievement procedure, the computational-time upper bound of metalevel actions at the low level is bounded by the number of intermediate nodes $N$ for each goal separately.
% Since it is assumed that all goal nodes are explored, a maximum of $N$ nodes can be explored in the low level policy for all goals.

The computational time
% complexity
upper bound of the hierarchical
% version of
BMPS is
\begin{align}
%   \textrm{Max Time} =  
   \sum_{i=0}^{M} \left[ (M- i) \cdot O (B) + O (B^{(M - i)}) \right] + \sum_{j=0}^{N}(N- j) \cdot O (B^{(N - j)})
   = O(M^2 \cdot B + B^{M}  + N^2 \cdot B^{N})
\end{align}

In the case of myopic approximation, the number of relevant nodes at each level is $1$.
% leading to the following simplification:
Therefore, the computational time upper bound to perform all metalevel actions and terminate is
\begin{align}
%   \textrm{Max Time} =
    \left( \sum_{i=0}^{M}(M- i) + \sum_{j=0}^{N}(N- j) \right) \cdot O (B)
  & = \left( \frac{M\cdot (M+1)}{2} + \frac{N\cdot (N+1)}{2}\right) \cdot O (B)  \\
  & = O((M^2 + N^2) \cdot B)
\end{align}

\section{Analysis of the speed-up achieved by the tree-contraction method} \label{sec:tree-contraction-speedup}
Table \ref{tab:contraction} shows an example computation of a single VPI feature \citep{callaway2017learning} computation. The computational speedup allows us to solve larger environments in the same amount of time and contributes to scaling the algorithm to more realistic problems.

\begin{table}[h]
\centering
\begin{tabular}{l|ccc}
\toprule
Environment size      & With tree contraction    & Without tree contraction    \\ 
\midrule
Branching (3,3,3), 27 nodes & 0.004s &   0.018s    \\ 
Branching (5,5,5), 125 nodes    & 0.007s &   12.27s  \\ 
Branching (6,6,6), 216 nodes & 0.013s & 328.22s \\
\bottomrule
\end{tabular}
\caption{Comparison of computation time for a single VPI feature calculation on different environment sizes. The environments follow a three-step branching structure where all nodes in the environment have 3, 5 or 6 child nodes depending on the environment size.}
\label{tab:contraction}
\end{table}